\documentclass[11pt]{article}

\usepackage[margin=1in]{geometry}
\usepackage[T1]{fontenc}
\usepackage[utf8]{inputenc}
\usepackage{lmodern}
\usepackage{microtype}
\usepackage{graphicx}
\usepackage{booktabs}
\usepackage{tabularx}
\usepackage{array}
\usepackage{float}
\usepackage[numbers,sort&compress]{natbib}
\usepackage[hidelinks]{hyperref}

\newcolumntype{Y}{>{\raggedright\arraybackslash}X}

\title{Prior Availability in Industrial Visual Sim-to-Real: A Review of CAD-Guided and CAD-Unavailable Regimes}
\author{
Chenxi Tao\\
George W. Woodruff School of Mechanical Engineering\\
Georgia Institute of Technology\\
\texttt{ctao40@gatech.edu}
\and
Seung-Kyum Choi\thanks{Corresponding author.}\\
George W. Woodruff School of Mechanical Engineering\\
Georgia Institute of Technology\\
\texttt{schoi@me.gatech.edu}
}
\date{}

\begin{document}
\maketitle

\begin{abstract}
Industrial visual sim-to-real is often described as transferring from
synthetic images to real images, but industrial deployment usually involves a
broader mismatch between available evidence and required decisions. A system
may be built from CAD renderings, simulated RGB-D observations, normal
reference images, synthetic defects, pretrained feature spaces, or language
prompts, yet deployed under different sensors, lighting, materials, fixtures,
calibration, production variation, and rare defect modes. This review
reframes industrial visual sim-to-real as a domain-gap problem organized by
prior availability. We distinguish CAD-available settings, where explicit
object geometry can support rendering, calibration, pose estimation,
segmentation, and test-time geometric verification; CAD-unavailable settings,
where geometry is replaced by normal-reference appearance, feature
distributions, teacher-student residuals, synthetic anomaly assumptions,
foundation features, or vision-language priors; and boundary-prior settings,
where approximate models, templates, reference views, or semantic
correspondences preserve only part of the CAD role. We make this distinction
operational with a rubric over source generation, correspondence, test-time
checking, and calibration support. This framing connects CAD-based detection
and 6D pose-estimation literature with industrial anomaly and
surface-inspection literature that is usually reviewed separately. To make the
taxonomy concrete, we use empirical anchors on T-LESS/BOP, MVTec AD, and VisA.
The anchors show that CAD render count alone does not close transfer;
source-distribution design, detector capacity, and small real calibration can
matter more. They also show that CAD at test time creates a distinct
verification channel through mask, pose, and depth consistency, whereas
CAD-unavailable inspection relies on calibrated normality and feature
deviation. The review therefore argues against a single cross-task leaderboard
and instead asks what prior grounds the deployment decision.
\end{abstract}


\vspace{0.4em}
\noindent\textbf{Project repository:}
\href{https://github.com/JacksonTao888/industrial-visual-sim2real-priors}{GitHub repository}

\section{Introduction}

Industrial visual recognition systems are increasingly expected to operate in
settings where large, fully labeled real deployment datasets are unavailable.
New products may enter a production line before many real images have been
collected. Defects may be rare by design, and therefore difficult to observe
before deployment. Pixel-level defect masks, 6D object poses, and detailed
inspection annotations can be expensive to produce. At the same time, factory
conditions can change with cameras, lighting, tooling, fixtures, batches,
surface materials, wear, and calibration. These pressures make industrial
vision a natural setting for sim-to-real and source-to-target transfer
problems. Broader sim-to-real and randomized-simulation reviews make the same
point in robotics: transfer performance depends on what aspects of the source
world are randomized, adapted, or preserved through deployment
\citep{muratore2022robotlearning,zhao2020simtorealsurvey}.

The phrase ``sim-to-real'' is sometimes used narrowly to mean transferring
from synthetic rendered images to real camera images. That case is important,
but it is too narrow for industrial vision. In practice, the source condition
may be a CAD-driven PBR rendering pipeline, simulated RGB-D observations, synthetic
defects, normal-reference images, pretrained visual features, benchmark
datasets, or language prompts. The target condition is the real deployment
environment: real sensors, optics, illumination, materials, backgrounds,
clutter, depth artifacts, object tolerances, production variation, rare real
defects, and operational reliability constraints. Industrial visual
sim-to-real is therefore a broader domain-gap problem: the information used to
design, train, validate, or configure the system differs from the condition in
which the system must make decisions.

This review argues that the most useful first split for this domain-gap
problem is whether CAD and related model priors are available. CAD
availability is not merely a dataset detail. It changes what kind of prior the
system can use, what kind of mismatch it faces, and what kinds of verification
are possible at test time. With CAD or a renderable object model, the system
can generate synthetic views, estimate pose, compare rendered masks or depth
against real observations, and use geometry to accept or reject visual
hypotheses. Without CAD, the system cannot directly render the target object
and test geometric alignment. It must instead infer whether an observation is
normal or abnormal from appearance references, feature distributions,
reconstruction residuals,
synthetic anomaly assumptions, or pretrained foundation-model representations.

\begin{figure}[H]
  \centering
  \includegraphics[width=\linewidth]{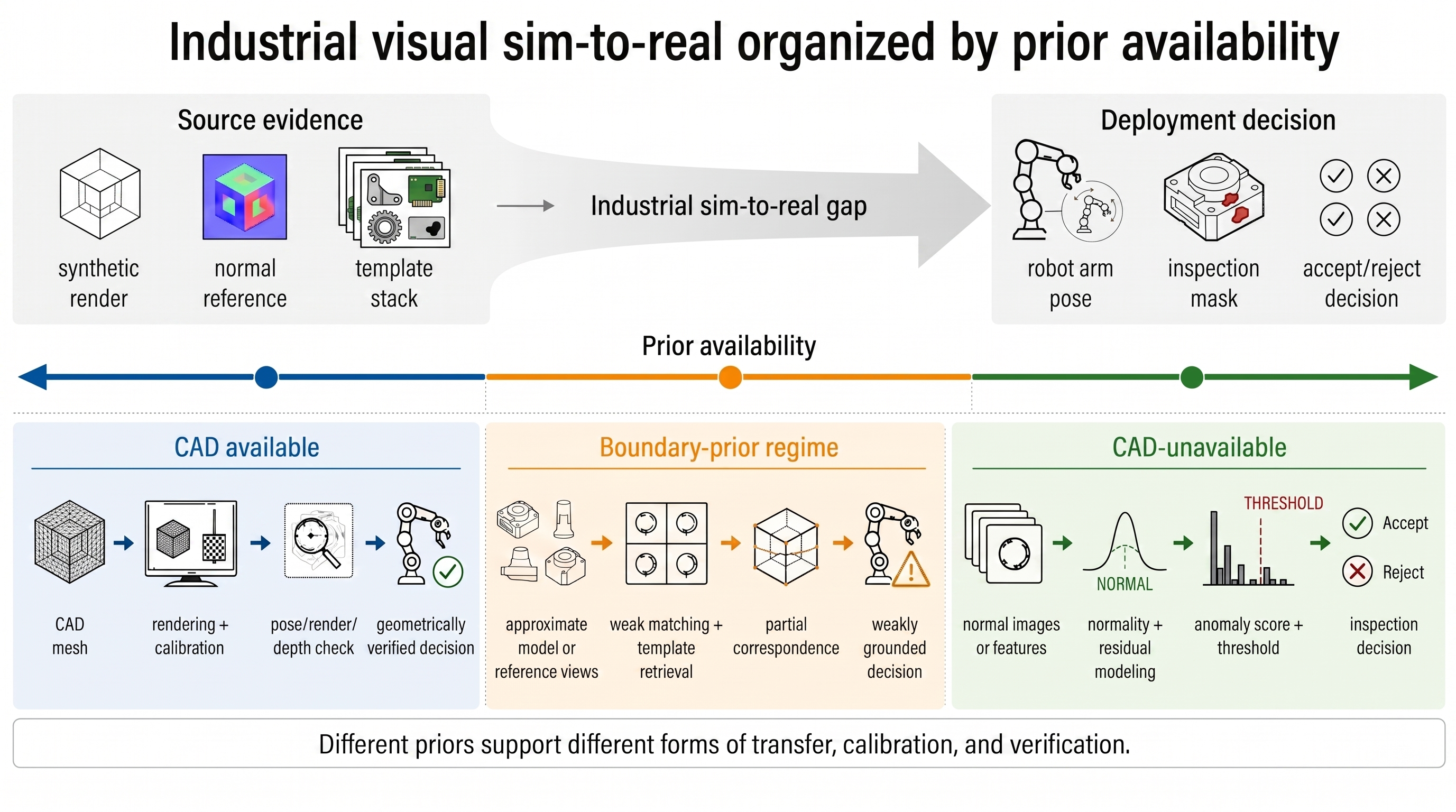}
  \caption{Prior-availability view of industrial visual sim-to-real. Source
  evidence may include explicit geometry, reference views, or learned feature
  priors, but these sources support different forms of transfer into factory
  deployment. CAD-available methods can use geometry for rendering,
  calibration, and test-time verification; boundary-prior methods preserve
  weaker correspondences or partial checks; CAD-unavailable inspection replaces
  explicit geometry with normal-reference modeling, feature residuals, anomaly
  scores, and thresholded decisions.}
  \label{fig:intro-prior-availability}
\end{figure}

Figure~\ref{fig:intro-prior-availability} summarizes this organizing axis.
It separates the available source evidence from the deployment decision that
the system must ultimately make, and places prior availability between the two.
The same deployment-domain gap can then be attacked through different
mechanisms: CAD-available methods can render, calibrate, align, and verify
against geometry; boundary-prior methods preserve weaker correspondence or
partial checking; and CAD-unavailable inspection relies on normality, feature
residuals, anomaly scores, and thresholds.

The thesis of the review is therefore simple: CAD makes the industrial
sim-to-real gap geometrically addressable; without CAD, the gap becomes
primarily an appearance and statistical generalization problem. This statement
is not meant to impose a rigid binary. Between full CAD availability and
complete CAD absence lies a boundary-prior regime, where systems use
approximate geometry, sparse references, templates, category-level shape,
foundation-model correspondences, or prompts. These methods blur the boundary,
but they still fit the review's central question: what prior is available, and
what kind of verification or calibration does it support? CAD availability
remains a useful organizing axis because it determines whether
explicit object geometry can be preserved through deployment and used as a
constraint.

The review connects two literatures that are often discussed separately.
CAD-available industrial vision includes known-object detection, 6D pose
estimation, robotic guidance, bin picking, assembly verification, CAD-based
segmentation, and render-and-compare verification. Datasets and benchmarks
such as T-LESS and BOP make this setting concrete by providing object models,
RGB-D observations, and 6D pose annotations for textureless industrial objects
\citep{hodan2017tless,hodan2018bop}. CAD-unavailable industrial inspection
includes surface defect detection, texture anomaly detection, normal-reference
inspection, and appearance-based quality control. Benchmarks such as MVTec AD
and VisA make this setting concrete by focusing on industrial anomaly
detection and segmentation without object-level CAD meshes
\citep{bergmann2019mvtec,zou2022visa}. The two branches differ not only in
datasets and metrics, but in the kind of prior used to confront deployment
variation.

The contribution of this review is to make that organizing axis explicit. It
first defines industrial visual sim-to-real as a deployment-domain-gap problem
rather than as a generic synthetic-data recipe. It then reviews
CAD-available and CAD-unavailable method families according to the
priors they use, while treating boundary-prior methods as intermediate cases
instead of forcing them into a separate top-level taxonomy. Finally, it uses
representative empirical anchors to show how the same question changes across
detection, pose, and anomaly inspection: what prior is available, and what
kind of transfer, calibration, or verification does that prior support?

\section{Industrial Vision Tasks and Deployment Domain Gaps}

Industrial vision is not a single task. It includes visual recognition,
measurement, and inspection functions used to support manufacturing,
logistics, quality control, and robot operation. Earlier surveys of industrial
vision systems describe applications ranging from inspection and measurement
to robot guidance and process monitoring \citep{malamas2003survey}. More
recent surveys of image-based quality control and smart-manufacturing
inspection similarly emphasize that industrial vision covers multiple
recognition and decision tasks rather than only defect classification
\citep{diers2023qualitycontrolsurvey,babic2021imageinspection,czimmermann2020defectsurvey}. For the
present review, the most relevant task families are surface defect detection,
industrial anomaly detection, object detection and part localization, 6D pose
estimation and robot guidance, assembly or completeness verification,
dimensional or geometric inspection, and peripheral identification tasks such
as OCR or traceability. This task diversity is one reason a single
sim-to-real recipe is unlikely to be adequate.

These task families encounter domain gaps because the development condition
rarely matches the deployment condition. In object detection and localization,
the system may be trained with synthetic or limited real labels, then deployed
under different backgrounds, object arrangements, lighting, occlusions, and
camera exposure. Industrial object detection surveys in smart manufacturing
make clear that detection is a practical manufacturing task, not simply a
generic benchmark problem \citep{ahmad2022objectdetection}. In robot guidance,
visual servoing, and pose-driven manipulation, the recognized object must also
be localized accurately enough to support physical action
\citep{perez2016robotguidance}. A small pose, calibration, or depth error can
therefore become a practical robot failure. In surface inspection and anomaly
detection, the system may have many normal images but few or no real defect
examples. The gap is then not only between synthetic and real images, but
between available assumptions about abnormality and the real defect mechanisms
that production eventually creates.

The sim-to-real domain gap exists for several recurring reasons. Real data are
costly, delayed, or unavailable before deployment. A manufacturer may need to
inspect a new part before enough real images exist; a defect class may be rare
because failures are undesirable; 6D poses, instance masks, and pixel-level
defect labels may require specialized annotation; and production cannot always
be interrupted to collect balanced training sets. Simulation and proxy data are
therefore attractive because they are controllable. CAD rendering can vary
pose, camera, and illumination. Domain randomization deliberately expands
source-domain variation so that a real image may appear as one sample from a
broader randomized distribution \citep{tobin2017domainrandomization}.
Synthetic anomaly generation can create training signals when real defects are
scarce. Normal-reference sets, foundation-model features, and language prompts
are also proxy sources, even though they are not simulations in the narrow
rendering sense.

However, the same controllability makes simulation incomplete. Rendering
pipelines approximate real optics, illumination, material reflectance,
transparency, specularity, background clutter, camera noise, blur, and depth
artifacts. CAD models may omit manufacturing tolerances, wear, deformation,
surface finish, and production variability. Synthetic defects may not match
the physical mechanisms that produce scratches, contamination, dents, missing
components, soldering defects, or logical assembly errors. Pretrained visual
features may transfer well enough for coarse recognition but not for dense
industrial localization or calibrated inspection thresholds. The result is a
source condition that is useful but incomplete.

Industrial tasks are also unusually sensitive to small spatial errors.
Classification errors matter, but many industrial decisions depend on precise
localization, alignment, or segmentation. A detector may need to find a small
textureless part in clutter. A pose estimator may need to guide a robot gripper
or verify assembly. A surface-inspection system may need to localize a tiny
defect while avoiding false alarms caused by normal texture variation. A
metrology or geometric-inspection system may need to decide whether a part is
within tolerance. In these settings, a domain shift that seems modest at the
image level can produce a large operational failure.

For these reasons, industrial visual sim-to-real is best defined as the problem
of designing visual recognition systems under source conditions that differ
from real deployment conditions, while using whatever priors are available to
reduce, calibrate, or verify that mismatch. The available prior is therefore
central. A CAD mesh, an approximate template, a few reference views, a normal
image memory bank, a synthetic defect generator, a pretrained visual backbone,
and a language prompt do not close the same gap in the same way. They encode
different assumptions about what remains stable between source and target, and
they support different kinds of test-time evidence. The task inventory
therefore leads back to Figure~\ref{fig:intro-prior-availability}: industrial
sim-to-real is not defined only by the visual task being solved, but also by
the prior that remains usable when source evidence meets deployment variation.
This is why the review moves from task families to prior availability before
separating method families.

\section{Prior Availability as the Organizing Axis}

This section unpacks the prior-availability axis introduced in
Figure~\ref{fig:intro-prior-availability}. The difference between
CAD-available and CAD-unavailable industrial vision is not simply a difference
in convenience. It changes the structure of the recognition problem and the
kind of evidence that can survive the move from source construction to factory
deployment. When CAD or a renderable model is available, the system has access
to an explicit description of object geometry. This geometry can be used before
deployment to synthesize training data, but it can also be used during
deployment to test visual hypotheses. When CAD is unavailable, the system lacks
this geometric reference and must replace it with appearance-based or
statistical priors.

\begin{figure}[H]
  \centering
  \includegraphics[width=\linewidth]{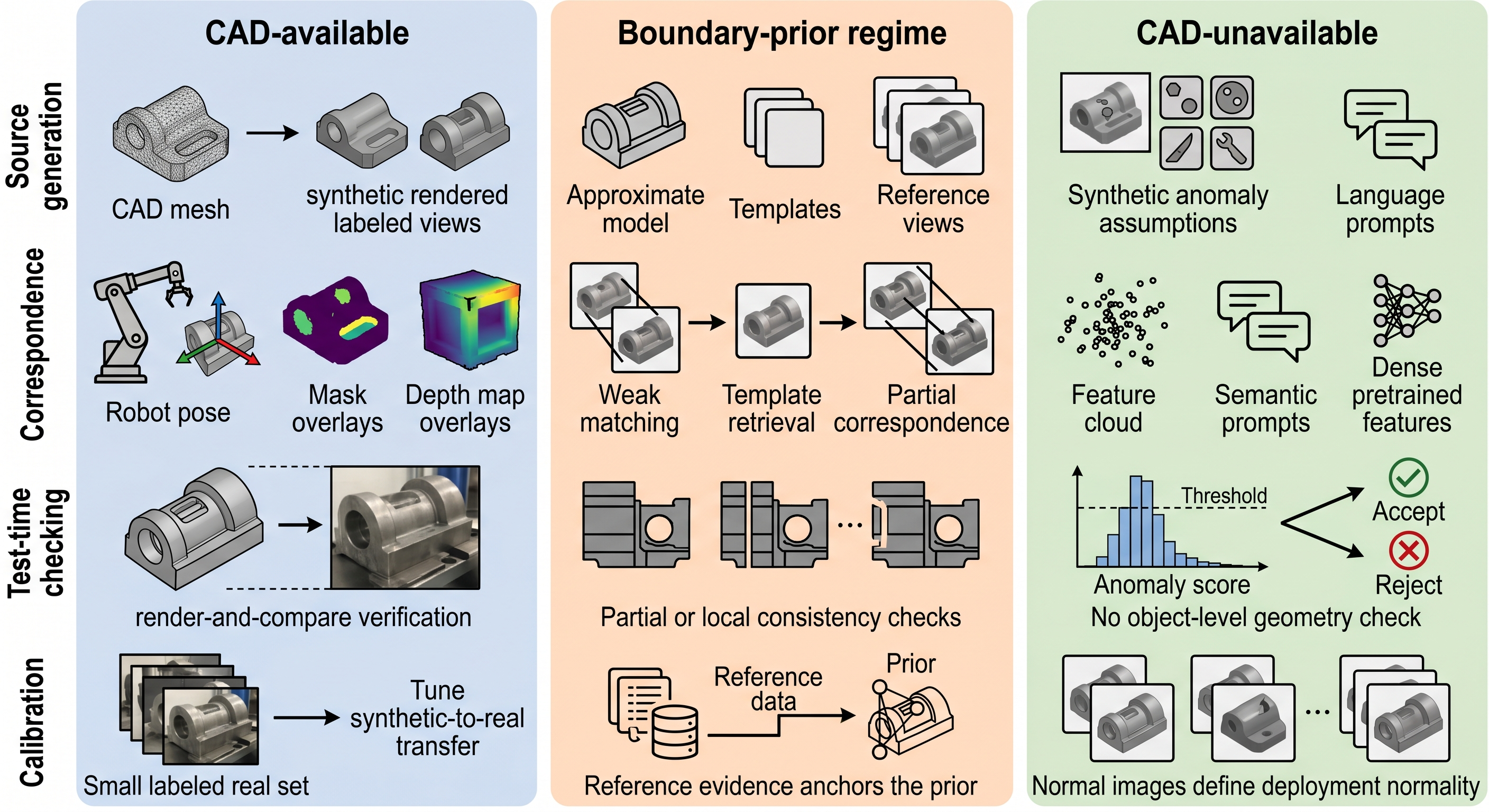}
  \caption{Mechanism view of the prior-availability distinction. CAD-available
  settings can use explicit geometry for source generation, correspondence,
  test-time checking, and calibration. Boundary-prior settings preserve only
  part of this evidence. CAD-unavailable inspection replaces object-level
  geometry with normal references, feature similarity, semantic priors,
  synthetic anomaly assumptions, or thresholded statistical evidence.}
  \label{fig:prior-mechanism-matrix}
\end{figure}

Figure~\ref{fig:prior-mechanism-matrix} turns this distinction into a mechanism
view. The same task family may require different kinds of evidence depending
on whether geometry is explicit, partial, or absent.

The mechanism view can be made operational with a simple prior-strength
rubric. Let a method be described by a vector over four channels: source
generation, correspondence, test-time checking, and calibration or decision
support. A strong CAD-available method scores high when the same object-level
geometry can generate labeled source data, establish pixel or pose
correspondence, and be rendered back into the real scene for verification. A
boundary-prior method preserves only part of this vector. A CAD-unavailable
method may score high on normal-reference calibration or semantic evidence,
but it has no object-level geometric checking channel. This rubric is not a
new universal benchmark; it is a reproducible way to state what kind of
evidence a method actually carries across the domain gap.

\begin{table}[H]
\centering
\caption{Operational rubric for prior strength. Each channel can be read as absent, weak, or strong depending on whether the method provides no usable evidence, partial or indirect evidence, or explicit object-level evidence for that role.}
\label{tab:prior-strength-rubric}
\footnotesize
\begin{tabularx}{\linewidth}{@{}p{0.18\linewidth}YYY@{}}
\toprule
Channel & Strong CAD-available evidence & Boundary-prior evidence & CAD-unavailable replacement \\
\midrule
Source generation & Renderable, scale-aware geometry produces labeled views, masks, depth, or poses. & Templates, approximate models, partial scans, or reference views support limited synthesis or retrieval. & Normal images, synthetic anomalies, prompts, or pretrained features define surrogate source evidence. \\
Correspondence & Pose, mask, depth, or surface coordinates map real observations to object geometry. & Weak matching, template retrieval, semantic localization, or partial correspondences are available. & Patch similarity, feature memory, language alignment, or residual maps provide non-geometric correspondence. \\
Test-time checking & A hypothesis can be rendered or projected back into the scene and checked geometrically. & Local consistency, reference-view agreement, or partial shape checks are possible. & Decisions rely on anomaly scores, thresholds, or semantic plausibility without object-level geometric verification. \\
Calibration and decision support & Real calibration can tune detector proposals, pose scoring, and geometric acceptance rules. & Small references or prompts tune only part of the pipeline. & Normal-reference coverage, false-alarm budgets, threshold stability, and category-specific normality define deployment behavior. \\
\bottomrule
\end{tabularx}
\end{table}

\subsection{CAD-Guided Geometry: Rendering and Verification}

In the CAD-available situation, the system may have object meshes,
camera intrinsics, depth observations, pose annotations, or renderable
templates. This situation is common for known industrial parts, bin picking,
robot guidance, 6D pose estimation, assembly verification, and model-based
inspection. Benchmarks such as T-LESS and BOP represent this situation because
they provide object models and pose-centric evaluation settings
\citep{hodan2017tless,hodan2018bop}. The domain gap appears when images or
geometric predictions derived from CAD do not match real factory observations.
Rendered images may differ in texture, illumination, clutter, object placement,
occlusion, and sensor artifacts. Yet CAD also provides a path that
appearance-only methods do not have: the method can align geometry to the
image, render a pose hypothesis, compare silhouettes or depth maps, and
enforce consistency.

This means CAD should not be understood only as a way to produce synthetic RGB
images. CAD has two roles. First, CAD is a pre-deployment renderer. It can
produce many labeled views, support PBR simulation, and enable domain
randomization. Rendering pipelines such as BlenderProc2 and BOP-style
synthetic generation are part of this role \citep{denninger2023blenderproc2}.
Second, CAD is a deployment-time geometric prior. Methods can use CAD for 6D
pose estimation, render-and-compare scoring, segmentation, and depth or mask
consistency. Classical model-based recognition and pose estimation already
used this idea \citep{drost2010ppf,hinterstoisser2012linemod}, and modern
methods extend it with learned features, render-and-compare architectures, and
foundation-model representations.

Recent CAD-available pose methods show this second role clearly. MegaPose uses
render-and-compare reasoning for novel-object 6D pose estimation
\citep{labbe2023megapose}. FoundationPose, FoundPose, SAM-6D, GigaPose, and
related methods push the setting toward stronger foundation features,
segmentation, zero-shot or unseen-object pose, and more robust template
matching
\citep{wen2024foundationpose,foundpose2024,lin2024sam6d,nguyen2024gigapose}.
The details differ, but the review-level point is shared: when CAD information
is available, geometry can remain active at test time. Closing the
sim-to-real gap is therefore not only a matter of making synthetic images look
more realistic; it is also a matter of using geometry to constrain or verify
recognition under real deployment conditions.

\subsection{CAD-Unavailable Inspection: Replacement Priors}

In the CAD-unavailable situation, useful object-level geometry is absent
or unavailable for the task. This situation is common in surface defect
detection, texture anomaly detection, and appearance-based quality inspection,
where the object geometry may be less important than surface condition,
material variation, or rare defects. It also occurs when CAD exists somewhere
in the engineering process but is not accessible, not aligned with the
inspection setup, not useful for the visible defect pattern, or too costly to
integrate. The absence of CAD does not mean the system has no prior
information. It may have normal images, a few reference examples,
category-level data, synthetic anomaly assumptions, pretrained visual
features, or language prompts. But none of these directly provides the same
object-level render-and-verify mechanism.

CAD-unavailable anomaly detection therefore treats the domain gap differently.
MVTec AD and VisA anchor the common benchmark setting: methods are given
normal training data and evaluated on their ability to classify or localize
anomalies in held-out real images \citep{bergmann2019mvtec,zou2022visa}.
Normal-reference memory methods such as PatchCore store dense normal features
and score deviations from that memory \citep{roth2022patchcore}.
Teacher-student methods such as EfficientAD use discrepancies between learned
representations as anomaly signals \citep{batzner2024efficientad}. Synthetic
anomaly methods such as DRAEM, SimpleNet, and SuperSimpleNet generate or
simulate abnormal patterns to create supervisory signals
\citep{zavrtanik2021draem,liu2023simplenet,rolih2024supersimplenet}.
Vision-language and foundation-feature methods such as WinCLIP and
AnomalyDINO replace explicit geometry with semantic prompts or dense
self-supervised visual features
\citep{jeong2023winclip,oquab2023dinov2,damm2025anomalydino}. These families
are different, but they all answer the same CAD-unavailable question: what can
stand in for geometry when geometry is not available?

\subsection{Boundary Priors Between CAD-Guided and CAD-Unavailable Regimes}

Between these endpoints is the boundary-prior regime introduced above. A
method may have an approximate mesh, a CAD template, a partial scan, a few
reference views, or foundation-feature correspondences. Such priors can support
weak matching, semantic localization, or limited verification, but they do not
provide the full geometric authority of a scaled, renderable CAD model. They
are therefore not treated as a third top-level taxonomy. Instead, they are
best understood by asking which part of the CAD role they preserve: source
generation, correspondence, calibration, or test-time checking.

\subsection{Real Data Across Prior Regimes}

The CAD-availability distinction also changes the role of real data. In CAD-available
detection and pose settings, a small amount of real labeled data can calibrate
a synthetic-trained detector, align render assumptions with real imagery, or
help tune proposal and verification thresholds. Real data support transfer
from CAD-rendered source images to real target observations. In CAD-unavailable
anomaly detection, real normal data often define the target distribution
itself. The system may not have real defect labels, but it can use normal references to
estimate what deployment normality looks like. Thus, small real data matter in
both situations, but the mechanism is different: supervised calibration in the
CAD branch, normal-reference modeling in the CAD-unavailable branch.

This distinction is the foundation for the rest of the review. In
CAD-available settings, the central questions are how to render
synthetic training data, how to randomize source variation, how to combine
synthetic and real calibration data, and how to use geometry at test time for
pose, segmentation, or verification. In CAD-unavailable settings, the
central questions are how to model normal appearance, how to calibrate anomaly
scores, how to synthesize useful defect assumptions, and how to adapt
pretrained visual or language-aligned features to industrial localization.
Boundary-prior methods are interpreted by the partial evidence they preserve
rather than by treating them as a separate leaderboard. The resulting flow is
therefore from the general industrial domain gap, to the prior that can reduce
or test that gap, and only then to the method families built around that prior.

This order matters because a method leaderboard would obscure the review's
main comparison. T-LESS detection or pose diagnostics, MVTec AD anomaly
localization, and VisA image-level anomaly classification are not directly
comparable; they expose different ways of confronting a domain gap under
different priors, labels, metrics, and deployment constraints. The literature
review therefore proceeds through two main branches. The CAD-available branch
asks how explicit geometry is used before deployment as a renderer and during
deployment as an alignment or verification prior. The CAD-unavailable branch
asks what replaces geometry when only normal references, synthetic anomaly
assumptions, pretrained features, or language-aligned priors are available.

Boundary-prior cases are introduced within those branches after the two
endpoints are clear. Approximate models, CAD templates, reference-light pose
methods, robust anomaly benchmarks, and LVLM-style inspection systems matter
because they show how the boundary between full CAD availability and complete
CAD absence is beginning to blur. They are best read as extensions of the
prior-availability argument, not as a separate third axis. The empirical
anchors later in the paper therefore serve as representative probes of these
families rather than as the source of the taxonomy.

\section{CAD-Guided Sim-to-Real Methods}

In CAD-available industrial vision, the sim-to-real problem is shaped by
an unusually strong prior: explicit object geometry. This prior does not remove
the domain gap. Rendered objects can still differ from real objects in
illumination, material response, background context, clutter, depth noise,
manufacturing tolerance, and sensor behavior. What CAD changes is the set of
possible responses to that gap. A CAD model can be used before deployment to
generate synthetic training data, and it can be used during deployment to make
visual hypotheses geometrically testable. The CAD-available literature is
therefore best read as two connected lines of work: CAD as a renderer and CAD
as a test-time geometric prior.

Table~\ref{tab:cad-available-summary} summarizes the main CAD-available
families by the role played by geometry. The table is not intended as a
leaderboard. It identifies what each family preserves from CAD and what kind
of deployment evidence that prior makes possible.

\begin{table}[H]
  \centering
  \footnotesize
  \setlength{\tabcolsep}{3pt}
  \renewcommand{\arraystretch}{1.18}
  \begin{tabularx}{\linewidth}{@{}>{\raggedright\arraybackslash}p{0.17\linewidth}YYY@{}}
    \toprule
    CAD role & Mechanism & Representative methods or anchors & Deployment evidence and limitation \\
    \midrule
    Synthetic renderer &
    Generate labeled source views under controlled pose, lighting, material,
    and background variation. &
    T-LESS/BOP, BlenderProc2, domain randomization
    \citep{hodan2017tless,hodan2018bop,denninger2023blenderproc2,tobin2017domainrandomization,sadeghi2017cad2rl,tremblay2018deepobjectpose,tremblay2018fallingthings}. &
    Supports scalable labels and calibration studies, but transfer depends on
    source coverage, material realism, and real-domain anchoring. \\
    Model-based recognition and learned pose &
    Match model, template, or learned geometric evidence to real observations
    to estimate identity, mask, or 6D pose. &
    PPF, LINEMOD, SSD-6D, PoseCNN, DenseFusion, PVNet, PVN3D
    \citep{drost2010ppf,hinterstoisser2012linemod,kehl2017ssd6d,xiang2018posecnn,wang2019densefusion,peng2019pvnet,he2020pvn3d}. &
    Enables object-level localization and pose reasoning, but can be weakened
    by occlusion, symmetry, calibration errors, and poor proposals. \\
    Render-and-compare verification &
    Keep CAD active at inference by rendering hypotheses and comparing masks,
    silhouettes, depth, or pose consistency. &
    MegaPose and FoundationPose-style CAD-aware pose systems
    \citep{labbe2023megapose,wen2024foundationpose}. &
    Turns deployment decisions into testable geometric hypotheses, but requires
    usable camera metadata, object identity, and refinement inputs. \\
    Foundation-assisted CAD matching &
    Use foundation features, segmentation, or template matching to make CAD
    correspondence more robust under appearance shift. &
    EPOS, SurfEmb, ZebraPose, FoundPose, SAM-6D, GigaPose, CNOS, MUSE
    \citep{hodan2020epos,haugaard2022surfemb,su2022zebrapose,foundpose2024,lin2024sam6d,nguyen2024gigapose,nguyen2023cnos,cho2025muse}. &
    Improves matching, segmentation, and proposal generation, but still depends
    on whether the prior supports true geometry or only weak correspondence. \\
    Boundary-prior variants &
    Use approximate meshes, partial scans, reference views, templates, or
    category-level shape when full CAD is absent. &
    Model-free and reference-light BOP settings, FreeZeV2, Pos3R
    \citep{nguyen2025bop2024,freezev2,deng2025pos3r}. &
    Preserves partial evidence for matching or calibration, but usually lacks
    the authority of full render-and-verify geometry. \\
    \bottomrule
  \end{tabularx}
  \caption{CAD-available method families organized by the role of explicit or
  partial geometry. Representative datasets, renderers, and methods are cited
  in the third column.}
  \label{tab:cad-available-summary}
\end{table}

\subsection{CAD as Source: Synthetic Labels, PBR, and Domain Randomization}

The most direct use of CAD is synthetic data generation. A mesh or 3D model can
be rendered under many object poses, camera viewpoints, lighting conditions,
backgrounds, materials, and occlusion patterns. For industrial parts, this is
especially attractive because object geometry is often known before a product
line is fully operational, while real labels may be costly or unavailable.
Datasets such as T-LESS and benchmark frameworks such as BOP make this setting
concrete: the object models and pose annotations enable synthetic views and
pose-centric evaluation for textureless rigid objects
\citep{hodan2017tless,hodan2018bop}. Rendering pipelines such as BlenderProc2
further formalize the production of photorealistic or procedurally varied
training data \citep{denninger2023blenderproc2}. Synthetic pose pipelines and
datasets such as DOPE and Falling Things show the same logic in robot-pose
settings: renderable object models can provide dense pose supervision before
real deployments supply abundant labels
\citep{tremblay2018deepobjectpose,tremblay2018fallingthings}.

Synthetic rendering addresses a practical bottleneck, but it creates a new
source condition whose coverage must be designed. A rendered image may provide
perfect labels, yet still be an incomplete proxy for factory deployment. The
gap can appear in lighting, reflectance, sensor noise, depth artifacts,
background fixtures, object placement, occlusion, and material wear. Domain
randomization responds by deliberately widening the source distribution, so
that real deployment images are more likely to fall inside the variation
encountered during training \citep{tobin2017domainrandomization}. In
industrial CAD rendering, this idea becomes a design question: which factors
should be randomized, and which should be modeled faithfully?
Related sim-to-real work also uses image adaptation, dynamics randomization,
and randomized-to-canonical mappings to reduce the source-to-target mismatch
\citep{shrivastava2017simgan,ganin2016dann,tzeng2017adda,hoffman2018cycada,peng2018dynamicsrandomization,james2019sim2sim}.

The important review point is that synthetic quantity is not the same as
synthetic coverage. Adding more rendered images may not reduce the gap if the
additional images repeat the same assumptions about lighting, materials,
backgrounds, or object context. Conversely, fewer images with better
distributional coverage or a small amount of real calibration data may transfer
more effectively. BOP challenge updates reflect this broader trend: modern
object pose systems increasingly depend on carefully designed synthetic
training, segmentation, detection, and pose pipelines rather than on rendering
volume alone \citep{sundermeyer2023bop2022}. For a review of industrial
sim-to-real, CAD-as-renderer methods should therefore be discussed as
distribution-design methods, not merely as a way to avoid manual annotation.

This framing also clarifies the role of real data in CAD-available pipelines.
When CAD is used as a renderer, real images can calibrate the synthetic source
domain. They may tune detector thresholds, correct systematic appearance
mismatch, adapt to a camera and lighting setup, or expose failure modes not
represented in the renderer. The source-to-target gap is not solved by
declaring the synthetic data realistic; it is reduced by combining CAD
coverage, randomized variation, and enough real evidence to anchor the
deployment condition.

\subsection{Model-Based Recognition and Learned 6D Pose}

CAD at test time has a longer history than modern foundation models. Classical
model-based recognition treated the object model as an active part of
inference. Point-pair features and LINEMOD-style methods are representative
historical anchors: they match local or template-based model evidence to real
observations and estimate object identity or pose under clutter
\citep{drost2010ppf,hinterstoisser2012linemod}. These approaches already
contain the central CAD-available idea: the object model is not only a training
resource, but a structure against which real observations can be compared.
Recent pose-estimation surveys similarly distinguish model-based,
model-free, RGB, RGB-D, instance-level, and category-level regimes, which
reinforces why CAD availability changes the evidence available to the system
\citep{he2021poseestimationreview,guan2024poseestimationreview}.

Deep 6D pose estimation kept this geometric objective while changing the
feature and learning machinery. BOP-style benchmarks encouraged methods that
combine synthetic training, object detection or segmentation, pose regression,
refinement, and geometric evaluation \citep{hodan2018bop}. RGB and RGB-D pose
methods such as SSD-6D, PoseCNN, DenseFusion, PVNet, DPOD, CDPN, PVN3D, EPOS,
CosyPose, and GDR-Net illustrate how the field moved from template and
coordinate prediction toward learned correspondences, keypoints, dense fusion,
refinement, and geometry-guided representations
\citep{kehl2017ssd6d,xiang2018posecnn,wang2019densefusion,peng2019pvnet,zakharov2019dpod,li2019cdpn,he2020pvn3d,hodan2020epos,labbe2020cosypose,wang2021gdrnet}.
These methods differ in their
architectures, but they share a CAD-available assumption: the target object is
specific enough that geometry can be represented, learned from, or checked.

For industrial sim-to-real, the relevance of this literature is not limited to
pose estimation as a standalone benchmark. Pose is a mechanism for reducing
deployment uncertainty. If a system can align a CAD model to a real image or
RGB-D observation, it can ask questions that an appearance-only detector
cannot ask: Does the rendered object silhouette agree with the observed mask?
Does the predicted depth agree with the measured depth? Is the proposed object
pose physically plausible under the camera model? These questions turn CAD
availability into a test-time verification path.

\subsection{Render-and-Compare as Test-Time Verification}

Render-and-compare methods make the deployment-time role of CAD especially
clear. Instead of treating recognition as a single feed-forward prediction,
they use the object model to generate hypotheses and compare rendered
evidence with the real observation. MegaPose is a representative modern method
in this family: it estimates 6D pose for novel objects through a
render-and-compare procedure that uses object models at inference time
\citep{labbe2023megapose}. The method is important for this review because it
embodies the claim that CAD can remain active after training, not merely serve
as a source of synthetic images.

This deployment-time use of CAD changes how the domain gap should be
understood. If a detector trained on synthetic or mixed data produces a
candidate box, a CAD-aware system can use pose, mask, silhouette, or depth
consistency to verify the candidate. The domain gap is then handled by a
combination of appearance transfer and geometric checking. Appearance transfer
helps the system propose objects under real lighting and clutter; geometry
helps judge whether those proposals are compatible with the known object
model. This is qualitatively different from a CAD-unavailable anomaly detector, which
can score feature deviation but cannot render the object to test alignment.

The same point explains why CAD-at-test-time methods remain limited in
practice. The geometry prior is powerful only if the pipeline supplies usable
inputs: reasonable object proposals, correct object identity, adequate camera
calibration, usable depth or mask information, and pose hypotheses that can be
refined rather than lost. Occlusion, textureless surfaces, symmetric objects,
transparent or reflective materials, and clutter can all weaken the practical
verification signal. CAD makes the gap geometrically addressable, but it does
not make the industrial scene simple.

These cases should be treated as stress tests of the verification channel, not
as incidental nuisances. Transparent or reflective surfaces break the
assumption that rendered appearance and sensed depth are reliable proxies for
object geometry. Severe symmetries make several poses equally plausible even
when the object is correctly detected. Multi-instance clutter creates
identity, occlusion, and proposal-association errors before geometric
verification begins. Real-time line inspection adds latency constraints that
can make a high-quality render-and-compare score impractical. A CAD-guided
paper should therefore report which part of the channel fails: proposal
generation, object identity, pose ambiguity, sensor geometry, mask agreement,
depth agreement, or decision latency.

\subsection{Foundation-Assisted CAD Matching and Segmentation}

Recent CAD-available work increasingly combines explicit geometry with
foundation-model features, segmentation, and template matching. FoundationPose
is a useful anchor because it unifies model-based and model-free 6D pose
estimation and tracking for novel objects \citep{wen2024foundationpose}.
FoundPose uses foundation features for unseen object pose estimation
\citep{foundpose2024}. SAM-6D brings segmentation-style foundation priors into
zero-shot 6D pose estimation \citep{lin2024sam6d,kirillov2023sam}. GigaPose
uses correspondence-oriented matching for fast and robust novel-object pose
\citep{nguyen2024gigapose}. FreeZeV2 and Pos3R extend the frontier toward
frozen foundation models and unseen-object pose estimation
\citep{freezev2,deng2025pos3r}. Earlier correspondence-heavy methods such as
SurfEmb and ZebraPose are useful bridges to this newer literature because they
already emphasize learned object-surface correspondence as a way to keep
geometry usable under image-domain variation
\citep{haugaard2022surfemb,su2022zebrapose}.

These methods should not be described as abandoning CAD. Rather, they show that
modern CAD-available systems increasingly use foundation features to make
geometry more usable under real appearance variation. The foundation model
helps match, segment, or represent the object across domains; the CAD or model
prior keeps the hypothesis grounded in object geometry. This combination is
important for industrial sim-to-real because it suggests a middle path between
pure rendering and pure appearance learning. Robust features help bridge the
appearance gap, while geometry remains available for alignment and
verification.

CAD-based 2D detection and segmentation form a related branch. Methods such as
CNOS and MUSE use CAD or model information to support novel-object
segmentation, detection, or similarity estimation without requiring a
traditional fully supervised detector for every target object
\citep{nguyen2023cnos,cho2025muse}. These methods matter because industrial
pipelines often need object proposals before pose estimation or verification
can begin. They also blur the line between detection, segmentation, and pose:
an object model can be used to find the object, segment it, estimate its pose,
or verify its consistency with the scene.

\subsection{Boundary Priors: Approximate Models and Reference-Light Pose}

CAD availability is a useful organizing axis, but it is not a perfect
binary. Recent BOP challenge framing explicitly includes both model-based and
model-free 6D pose estimation \citep{nguyen2025bop2024}. In practice, a system
may have an approximate CAD model, a partial scan, a few reference images, a
category-level shape prior, a CAD template, or a foundation-model
representation rather than a precise object mesh. These methods occupy a
boundary-prior regime between full CAD availability and complete CAD absence:
they provide more structure than ordinary CAD-unavailable inspection, but less
geometric authority than a scaled, renderable object model.

For the present review, this regime is treated as a boundary of the main
CAD-availability distinction rather than as a third primary taxonomy. The relevant
question is whether the available prior supports explicit geometric
verification, weak correspondence, semantic localization, or only calibration.
A full mesh with known scale and camera intrinsics supports render-and-compare
reasoning directly. A few reference views may support matching but not complete
depth or silhouette verification. A CAD template may help 2D localization
without providing full pose verification. A language prompt may identify a
semantic target but provides no object-specific geometry. These differences
matter for deployment because they determine what kind of domain gap can be
checked at test time.

\subsection{Synthesis of the CAD-Guided Branch}

The CAD-available literature supports a central conclusion: CAD makes
industrial sim-to-real geometrically addressable, but only when the pipeline
uses geometry in the right way. As a renderer, CAD can produce scalable labeled
source data, but transfer depends on distributional coverage, domain
randomization, and real calibration. As a test-time prior, CAD can support
pose estimation, render-and-compare, segmentation, mask consistency, and depth
verification, but these signals depend on proposal quality, calibration,
occlusion handling, and robust feature matching. Modern CAD-aware methods that
use foundation features do not replace this logic; they strengthen it by
combining transferable features with explicit geometry.

This sets up the contrast with CAD-unavailable inspection. Without CAD,
there is no mesh to render, no object-specific silhouette to verify, and no
pose hypothesis to check against depth. The next section therefore asks what
can substitute for explicit geometry when industrial vision must rely on
normal-reference memory, appearance statistics, synthetic anomaly assumptions,
or foundation-model features.

\section{CAD-Unavailable Industrial Inspection}

When CAD information is unavailable, industrial sim-to-real changes
character. The system can no longer render the target object, align a mesh to
the image, or verify a hypothesis through object-specific silhouette or depth
consistency. This does not mean the system has no prior. It may have normal
training images, a few reference examples, category-level data, synthetic
defect assumptions, pretrained visual features, language prompts, or
unlabeled test images. The CAD-unavailable literature is therefore not a literature of
``no information.'' It is a literature of replacement priors: what can
substitute for explicit geometry when the inspection problem is dominated by
appearance, texture, material variation, and rare defects?

This branch also sits within the broader anomaly-detection literature, where
surveys have long emphasized the dependence of anomaly scores on assumptions
about normality, rarity, feature representation, and decision thresholds
\citep{chandola2009anomaly,pang2021deepanom,ruff2021unifying}. Industrial
anomaly surveys further separate reconstruction, feature-embedding,
teacher-student, synthetic-anomaly, and foundation-model directions
\citep{jin2022surfacedefect,tao2022unsupervisedanomalysurvey,liu2024deepindustrialsurvey}. The industrial
case is narrower but more demanding: the anomaly score must be converted into
a reliable quality-control decision under real process variation.

Industrial anomaly detection benchmarks make this setting concrete. MVTec AD
defines a canonical unsupervised industrial anomaly setting in which models
learn from normal images and are evaluated on anomaly classification and
localization \citep{bergmann2019mvtec}. VisA expands this setting with more
varied object and defect categories, including challenging PCB and
multi-instance scenes \citep{zou2022visa}. These benchmarks do not provide
object meshes for render-and-compare verification. Instead, they ask whether a
method can define a useful reference for normality and detect departures from
that reference under real image variation.

Table~\ref{tab:cad-unavailable-summary} summarizes the replacement-prior
families that structure the CAD-unavailable branch. Here the key comparison is not
which model is best on a benchmark, but what evidence substitutes for
object-level geometry and what deployment risk remains.

\begin{table}[H]
  \centering
  \footnotesize
  \setlength{\tabcolsep}{3pt}
  \renewcommand{\arraystretch}{1.18}
  \begin{tabularx}{\linewidth}{@{}>{\raggedright\arraybackslash}p{0.17\linewidth}YYY@{}}
    \toprule
    Replacement prior & Mechanism & Representative methods or anchors & Deployment evidence and limitation \\
    \midrule
    Normal-reference memory &
    Store or model dense normal features and score distance from the normal
    reference distribution. &
    Deep SVDD, SPADE, PaDiM, Patch SVDD, PatchCore, DifferNet, FastFlow,
    CFLOW-AD
    \citep{ruff2018deepsvdd,cohen2020spade,defard2021padim,yi2021patchsvdd,roth2022patchcore,rudolph2021differnet,yu2021fastflow,gudovskiy2022cflow}. &
    Works well when normal data cover legitimate variation, but thresholds and
    masks can drift under lighting, material, or process changes. \\
    Teacher-student residuals &
    Detect abnormal regions through discrepancies between teacher and student
    representations learned from normal behavior. &
    Uninformed Students, ST-FPM, MKD, Reverse Distillation, EfficientAD
    \citep{bergmann2020uninformed,wang2021stfpm,salehi2021mkd,deng2022reversedistillation,batzner2024efficientad}. &
    Provides dense and efficient residual evidence, but residual magnitude is
    only meaningful after calibration to deployment normality. \\
    Synthetic-anomaly supervision &
    Create abnormality by corrupting normal images or features when real defect
    labels are scarce. &
    CutPaste, DRAEM, NSA, SimpleNet, SuperSimpleNet, DeSTSeg
    \citep{li2021cutpaste,zavrtanik2021draem,schlueter2021nsa,liu2023simplenet,rolih2024supersimplenet,zhang2023destseg}. &
    Supplies training signals for rare defects, but can teach shortcut
    boundaries if synthetic defects do not resemble real process failures. \\
    Vision-language weak priors &
    Use language-aligned features or prompts to compare normal and abnormal
    semantic states. &
    CLIP, WinCLIP, AnomalyCLIP, APRIL-GAN, AdaCLIP, FiLo, MuSc
    \citep{radford2021clip,jeong2023winclip,zhou2023anomalyclip,chen2023aprilgan,cao2024adaclip,gu2024filo,li2024musc}. &
    Reduces object-specific supervision, but broad semantics may miss small,
    texture-level, or process-specific defects. \\
    Dense foundation features &
    Use pretrained patch features or few-shot references as transferable visual
    similarity priors. &
    DINOv2, AnomalyDINO, UniVAD
    \citep{oquab2023dinov2,damm2025anomalydino,gu2025univad}. &
    Preserves spatial comparison without CAD, but still requires reliable
    score calibration and cannot verify object-specific geometry. \\
    Robust, generative, and LVLM boundaries &
    Stress-test CAD-unavailable priors under logical anomalies, domain shifts, generated
    defects, or multimodal reasoning. &
    MVTec LOCO AD, MVTec 3D-AD, Real-IAD, RobustAD, AnomalyDiffusion,
    AnomalyGPT, AD-Copilot
    \citep{bergmann2022mvtecloco,bergmann2022mvtec3d,wang2024realiad,pemula2025robustad,hu2023anomalydiffusion,gu2024anomalygpt,jiang2026adcopilot}. &
    Expands inspection realism, but generated or language-based evidence must
    still be grounded in calibrated visual decisions. \\
    \bottomrule
  \end{tabularx}
  \caption{CAD-unavailable inspection families organized by the prior that
  replaces explicit object geometry. Representative datasets and methods are
  cited in the third column.}
  \label{tab:cad-unavailable-summary}
\end{table}

\subsection{Normal-Reference Memory and Feature Distributions}

The most direct CAD-unavailable prior is a collection of normal examples. Rather than
rendering a known object under hypothetical poses, normal-reference methods
estimate what normal appearance looks like in feature space. PaDiM models
patch-level feature distributions for anomaly detection and localization
\citep{defard2021padim}. Deep SVDD and Patch SVDD provide one-class
objectives at image or patch level \citep{ruff2018deepsvdd,yi2021patchsvdd}.
SPADE and PatchCore store representative normal features or correspondences
and score test patches by their distance from normal neighbors
\citep{cohen2020spade,roth2022patchcore}. Flow-based methods such as
DifferNet, FastFlow, and CFLOW-AD model feature likelihoods through
normalizing flows
\citep{rudolph2021differnet,yu2021fastflow,gudovskiy2022cflow}. These methods
differ technically, but they share the same review-level premise: without
object geometry, deployment normality must be approximated from real or
representative normal appearance.

Reconstruction-based baselines are part of the same historical thread. For
example, SSIM-based autoencoder defect segmentation asks whether a test image
can be reconstructed as normal appearance \citep{bergmann2019ssim}. Modern
feature-memory and flow methods often outperform such simple reconstruction
criteria, but the underlying prior remains similar: normality is defined by
what the model can represent or reconstruct from normal evidence.

This family is important because it makes the role of real data explicit. In a
CAD-available detector, a small real set may calibrate transfer from rendered
source images. In a CAD-unavailable anomaly detector, normal real images define the
reference distribution itself. The method does not need real defect examples,
but it does need the normal set to cover legitimate variation in surfaces,
materials, lighting, camera exposure, background, and production batches. The
sim-to-real gap appears when the learned normal distribution is too narrow or
when benchmark normality fails to represent deployment normality.

Patch-memory and feature-distribution methods are also a useful counterweight
to overly broad claims about foundation models. They are simple relative to
large vision-language systems, but they remain strong because they align well
with the industrial assumption that normal production data are easier to
collect than defect data. Their limitation is also clear: they usually detect
statistical deviation rather than explain defect mechanism, assembly logic, or
semantic abnormality.

\subsection{Teacher-Student Residuals and Efficient Inspection}

Teacher-student methods replace explicit geometry with representation
consistency. The system learns how a student model should reproduce or match
features from a teacher on normal data; abnormal regions are detected through
residuals or mismatches. Uninformed Students is a useful early anchor for the
industrial version of this idea \citep{bergmann2020uninformed}.
Student-teacher feature pyramid matching and multiresolution knowledge
distillation further show how feature hierarchies can supply dense residual
signals \citep{wang2021stfpm,salehi2021mkd}. Reverse Distillation uses a
one-class embedding and reverse distillation to detect deviations from normal
representations \citep{deng2022reversedistillation}. EfficientAD is a
particularly relevant industrial anchor because it emphasizes accurate visual
anomaly detection at millisecond-level latency \citep{batzner2024efficientad}.

For the sim-to-real framing, teacher-student methods show another way to
substitute for CAD geometry. The method cannot ask whether a rendered object
matches the image, but it can ask whether the representation behaves as it did
on normal data. This makes the anomaly score a residual against learned normal
behavior. The appeal is practical: the approach can be fast, dense, and trained
without defect labels. The risk is that residuals are only as meaningful as
the normal training distribution and the teacher representation. Changes in
lighting, material finish, camera settings, or background can look abnormal
even when they are acceptable deployment variation.

This is why CAD-unavailable inspection should not be evaluated only by image-level
AUROC. Industrial deployment often needs calibrated masks, controlled false
alarms, and stable thresholds under process drift. A method may rank abnormal
images well while producing unstable dense masks or thresholds. Teacher-student
methods therefore belong in the review not only as a model family, but as a
reminder that CAD-unavailable sim-to-real requires calibration of residual signals, not
just feature extraction.

\subsection{Synthetic-Anomaly Self-Supervision}

Synthetic-anomaly methods address a different CAD-unavailable bottleneck: real defects
are scarce, but the model still needs to learn abnormality. Instead of
rendering a CAD object, these methods synthesize defects on real normal
images or features. Earlier generative anomaly-detection work such as AnoGAN
and GANomaly helped establish reconstruction or adversarial deviation as a
usable abnormality signal, even though these methods were not designed
specifically for modern industrial inspection
\citep{schlegl2017anogan,akcay2018ganomaly}. CutPaste creates
self-supervised anomaly cues through cut-and-paste transformations
\citep{li2021cutpaste}. DRAEM combines
reconstruction and discriminative learning using synthetic anomalous patterns
\citep{zavrtanik2021draem}. Natural Synthetic Anomalies, SimpleNet,
SuperSimpleNet, and DeSTSeg represent related ways of using artificial
defects, feature perturbations, or denoising student-teacher mechanisms to
train localization and discrimination
\citep{schlueter2021nsa,liu2023simplenet,rolih2024supersimplenet,zhang2023destseg}.

This family is the CAD-unavailable analogue of synthetic data generation, but the
meaning of ``synthetic'' is different. In CAD-available recognition, synthetic
data usually come from rendering known object geometry. In CAD-unavailable anomaly
detection, synthetic data often come from corrupting normal images or features
to mimic unknown defects. The core question is therefore not whether the
synthetic anomaly is visually plausible in the abstract. It is whether the
synthetic anomaly distribution teaches a boundary that transfers to real
defect mechanisms.

The transfer risk is substantial. Real industrial defects can be caused by
scratches, dents, contamination, missing components, soldering failures,
logical assembly errors, material inconsistencies, or process-specific
mechanisms. A synthetic corruption may encourage useful localization behavior,
but it may also create shortcuts that do not correspond to deployment defects.
For this reason, synthetic-anomaly methods should be discussed as an important
source-prior family, not as a guaranteed replacement for defect data.

\subsection{Vision-Language Priors for Zero- and Few-Shot Inspection}

Vision-language models introduce another CAD-unavailable prior: semantic alignment
between images and language. CLIP provides a general visual-language feature
space learned from large-scale image-text supervision
\citep{radford2021clip}. WinCLIP adapts this idea to zero- and few-shot
anomaly classification and segmentation, making it a natural anchor for
language-aligned industrial inspection \citep{jeong2023winclip}. Subsequent
methods such as AnomalyCLIP, APRIL-GAN, AdaCLIP, FiLo, and MuSc explore
object-agnostic prompts, challenge-specific zero/few-shot pipelines,
learnable prompt adaptation, fine-grained descriptions, and mutual scoring of
unlabeled images
\citep{zhou2023anomalyclip,chen2023aprilgan,cao2024adaclip,gu2024filo,li2024musc}.

The attraction of VLM-based anomaly detection is clear. If language prompts
can describe normal and abnormal states, then a system may need fewer
object-specific references and may generalize across categories more easily.
This is especially tempting for industrial deployments where new products or
defects appear before a full training set exists. In the review's terms,
language becomes a weak prior that partially replaces object geometry and
normal-reference memory.

The limitation is equally important. A language-aligned feature space is not
automatically a dense industrial defect localizer. Many defects are small,
texture-level, material-specific, or process-specific, and they may not align
well with broad semantic descriptions. A prompt can say ``damaged'' or
``scratched,'' but deployment requires deciding whether a particular small
region under a particular camera and material is unacceptable. VLM methods
therefore belong in the CAD-unavailable section as promising weak-prior approaches, but
they should not be treated as having solved dense industrial sim-to-real.

\subsection{Dense Visual Foundation Features and Few-Shot Reference}

Dense self-supervised visual features offer a different foundation-model prior
from language alignment. DINOv2 provides robust visual representations learned
without language supervision \citep{oquab2023dinov2}. AnomalyDINO adapts
DINOv2-style patch features for few-shot anomaly detection
\citep{damm2025anomalydino}. UniVAD further reflects the trend toward
training-free or few-shot unified visual anomaly detection
\citep{gu2025univad}. These methods are important because they address a
practical middle ground: the system may lack CAD and defect labels, but it may
have a few normal reference images and a strong pretrained visual backbone.

This branch should be distinguished from CLIP-style anomaly detection. CLIP
provides language-aligned semantics; DINOv2-style features provide dense
visual correspondence and patch-level representation. For industrial
inspection, dense visual features may be more naturally suited to localization
because they preserve spatial structure and texture-level similarity. They can
therefore behave more like a learned normal-reference representation than like
a prompt-driven semantic classifier.

The open question is calibration. Dense foundation features may transfer well
across categories, but industrial anomaly detection still requires converting
feature distances or maps into reliable image-level decisions and pixel-level
masks. Larger backbones or stronger pretraining do not automatically solve
threshold selection, false-alarm control, or robustness to production shifts.
This keeps the CAD-unavailable problem fundamentally different from CAD-at-test-time
geometry: foundation features can compare appearances, but they cannot verify
object-specific shape or pose unless coupled with some model or reference
structure.

\subsection{Robustness, Logical Anomalies, and Deployment Realism}

The CAD-unavailable literature has also begun to expose the limits of standard anomaly
benchmarks. MVTec LOCO AD extends the setting beyond local texture defects to
logical constraints and structural anomalies \citep{bergmann2022mvtecloco}.
MVTec 3D-AD introduces 3D industrial anomaly detection and localization,
highlighting cases where geometry matters even without object-level CAD
\citep{bergmann2022mvtec3d}. Real-IAD provides a larger multi-view industrial
anomaly benchmark \citep{wang2024realiad}. RobustAD targets robustness under
real-world domain shifts \citep{pemula2025robustad}. MVTec AD 2 further
emphasizes advanced scenarios for unsupervised anomaly detection
\citep{hecklerkram2026mvtecad2}.

These datasets are important for the review because they prevent an overly
narrow interpretation of CAD-unavailable success. A method that performs well on a
curated benchmark may still fail under new lighting, new camera placement,
changed materials, unseen normal variation, logical assembly constraints, or
multi-view inspection requirements. In CAD-available settings, geometry can
sometimes be used to reject physically inconsistent hypotheses. In CAD-unavailable
settings, robustness must be achieved through reference coverage, feature
stability, calibration, and benchmark realism.

Three-dimensional and cross-modal inspection make this point especially clear.
Depth maps and point clouds can detect dents, missing material, deformation,
or assembly errors that are weakly visible in RGB, but they do not automatically
restore the full CAD role. Without a scaled object model, a depth-based anomaly
detector can compare local geometry to normal references, fuse photometric and
geometric residuals, or check multi-view consistency, but it usually cannot ask
whether the whole part matches a known design. Such methods are therefore
boundary cases within the CAD-unavailable branch: they add geometric evidence,
yet the evidence is empirical and reference-based rather than object-model
verified.

The deployment lesson is that anomaly detection is not only about ranking
anomaly scores. It is about deciding what evidence is sufficient to stop a
line, reject a product, or request human review. Thresholds, false positives,
mask quality, image-level decisions, and explanation all matter. This is why
robustness datasets and logical anomaly benchmarks should appear in the
review even if they are not part of the current empirical anchor suite.

\subsection{Generative and LVLM Boundary Directions}

Generative methods and large vision-language models define an emerging
boundary for CAD-unavailable inspection. Diffusion-based and generative
defect synthesis methods such as AnomalyDiffusion, SeaS, AnoGen, and
GroundingAnomaly attempt to create realistic defect images, masks, or
spatially grounded anomalies when real defects are scarce
\citep{hu2023anomalydiffusion,seas2024,gui2025anogen,liu2026groundinganomaly}.
These methods extend the synthetic-anomaly idea toward more realistic and
controllable data generation.

LVLM-based inspection methods move in a different direction. AnomalyGPT,
IADGPT, IAD-GPT, and AD-Copilot use large multimodal models for anomaly
detection, localization, reasoning, or visual in-context comparison
\citep{gu2024anomalygpt,iadgpt2025,iadgpt2025visualknowledge,jiang2026adcopilot}.
Their promise is not only to score anomalies, but also to compare examples,
describe defects, support interactive inspection, or provide reasoning that a
human operator can use.

For this review, these methods should be treated as boundary and future-work
directions rather than a third main taxonomy. They are important because they
weaken the requirement for object-specific data, but they do not remove the
core deployment question: what evidence grounds the decision? A generated
defect is useful only if it resembles real process failures. An LVLM
explanation is useful only if it is tied to reliable visual evidence and
calibrated inspection behavior. Without CAD, even advanced generative or
language-based systems still need some substitute for explicit geometric
verification.

\subsection{Synthesis of the CAD-Unavailable Branch}

The CAD-unavailable literature supports the second half of the review's
thesis. Without object-level geometry, industrial sim-to-real becomes a problem
of defining and calibrating appearance-based priors. Normal-reference memory
and feature distributions estimate what deployment normality looks like.
Teacher-student methods detect residuals from learned normal behavior.
Synthetic-anomaly methods create substitute abnormality when real defects are
rare. Vision-language methods introduce semantic weak priors. Dense visual
foundation features provide transferable patch representations. Robustness
benchmarks, generative synthesis, and LVLM methods extend the boundary toward
more realistic and interactive inspection.

These families should be compared by the role of their prior, not by a single
cross-task ranking. In the CAD-available branch, explicit geometry makes
alignment and render verification possible. In the CAD-unavailable branch, methods must
instead ask whether a test image is consistent with normal examples,
appearance statistics, synthetic defect assumptions, or pretrained feature
spaces. This contrast is the reason the empirical part of the paper should be
organized as representative anchors for the two situations rather than as one
leaderboard over unrelated industrial vision tasks.

\section{Empirical Anchors Across Prior-Availability Regimes}
\label{sec:empirical-anchors}

The empirical component is not an exhaustive benchmark study. It is a set of
representative anchors that makes the review taxonomy concrete. When CAD and
related model priors are available, the central question is how explicit
geometry can reduce, constrain, or diagnose the synthetic-to-real gap. When
CAD priors are unavailable, the question changes to which appearance,
normality, or foundation-model priors remain reliable enough for inspection.
The two branches therefore use
different tasks, datasets, and metrics. Their numerical results should not be
read as a single leaderboard; they are used to compare the mechanisms by which
prior availability changes industrial sim-to-real transfer.

Boundary-prior methods are treated here as interpretive boundary cases rather
than as a third empirical branch, because their weak priors vary from
approximate geometry to sparse references, templates, prompts, and
foundation-model correspondences.

The CAD-available branch uses T-LESS/BOP as a textureless industrial-object
setting with CAD models, RGB-D observations, camera metadata, and pose
annotations \citep{hodan2017tless,hodan2018bop}. It separates two roles that
are often collapsed in discussions of synthetic data: CAD as a pre-deployment
renderer and CAD as a test-time geometric prior. The CAD-unavailable branch
uses MVTec AD and VisA as industrial anomaly-detection settings without
object-level CAD meshes \citep{bergmann2019mvtec,zou2022visa}. It asks what
replaces geometry in practice: normal-reference memory, teacher-student
residuals, synthetic anomaly assumptions, language-aligned priors, and dense
visual foundation features.

\subsection{Experimental Design and Metric Logic}

Public datasets are used because this section supports a review argument rather
than a private deployment case study. The datasets are therefore chosen for the
kind of prior they expose. T-LESS/BOP anchors the CAD-available branch:
its object meshes, RGB-D real images, camera metadata, and 6D pose annotations
allow CAD to be examined both as a synthetic PBR renderer before deployment and
as render-and-compare geometry during deployment. The detector runs train on
synthetic PBR images and evaluate on held-out real Primesense images, so their
detection metrics are interpreted as known-object synthetic-to-real transfer
diagnostics rather than as a generic detector ranking.

MVTec AD and VisA anchor the CAD-unavailable branch. MVTec AD connects
the experiments to the canonical unsupervised industrial-inspection literature,
whereas VisA provides a larger and more varied benchmark used by recent
zero-shot, few-shot, and foundation-feature anomaly methods. Both datasets
provide normal training images and real anomalous test images with image-level
and pixel-level labels, but neither provides object-level CAD models for pose
or render verification. They therefore make the CAD-unavailable appearance-prior problem
concrete.

The anchors are selected by tracing the preceding literature review back to
representative mechanisms. The CAD-as-renderer block tests the claim made by
synthetic-rendering and domain-randomization work: CAD can provide labels, but
transfer depends on source distribution, model capacity, and real calibration
rather than render count alone. The CAD-at-test-time block tests the second CAD
role emphasized by pose and render-and-compare methods: geometry can remain
active at deployment as an alignment or verification signal. The CAD-unavailable block
then selects runnable representatives from the main replacement-prior families
reviewed above: normal-reference memory, teacher-student residuals, zero-shot
language-aligned priors, and dense visual foundation features. The goal is
explanatory coverage of method families, not an exhaustive search for the best
implementation in each family. The boundary-prior regime is not benchmarked as
a separate block because it is not defined by a shared task or metric; it is
defined by how much weak geometry, reference evidence, semantic localization,
or calibration support remains when full CAD is unavailable.

The metrics follow the mechanism being tested. For renderer-only detection,
mAP$_{50}$ and mAP$_{50:95}$ are used because the question is whether a
detector trained from CAD-rendered source images can localize known real
objects under standard detection criteria. mAP$_{50}$ captures coarse
object-level transfer, whereas mAP$_{50:95}$ is stricter about localization
quality across IoU thresholds. Other detector diagnostics, such as per-class
AP, recall at fixed confidence, false positives per image, precision-recall
curves, and confusion matrices, are useful and retained in the appendix, but
they are secondary to the review-level transfer question. F1 is not used as
the primary CAD-as-renderer metric because it depends on a selected confidence
threshold, whereas mAP summarizes the detector's precision-recall behavior
over operating points and is the standard detection measure for this setting.
For CAD-at-test-time geometry, mask IoU, visible coverage, ROI recall, and
proposal AUROC are used
because the question is no longer only detection accuracy; it is whether CAD
rendering, pose, and depth consistency can verify a real visual hypothesis.
Here too, F1 would require defining a binary acceptance threshold for each
verification score, while the anchor is intended to measure geometric overlap,
proposal coverage, and separability before committing to deployment-specific
costs.
For CAD-unavailable anomaly detection, image AUROC captures image-level ranking, pixel
AUROC captures localization ranking, and F1 summarizes thresholded operating
behavior. Metrics such as PRO, average precision, calibration error, latency,
or cost-weighted false-alarm rate would be appropriate in a deployment-focused
benchmark, but the present section uses AUROC/F1 because they are common across
MVTec AD and VisA and directly expose the distinction between ranking and
thresholded mask behavior.

This choice also defines the limits of the empirical anchors. They are not a
meta-analysis and they do not claim to estimate a single effect size across all
industrial tasks. A consolidated follow-up study would hold the target
deployment set fixed and vary source coverage, source volume, model capacity,
real calibration budget, verification channel, and controlled domain shifts.
The review's anchors are designed to motivate that protocol: they show which
variables must be separated before claims such as ``more rendering closes the
gap'' or ``geometry improves confidence'' can be interpreted quantitatively.

\begin{table}[H]
\centering
\caption{Controlled protocol implied by the review taxonomy. The rows identify the experimental factors that should be separated before comparing CAD-guided, boundary-prior, and CAD-unavailable methods.}
\label{tab:controlled-protocol}
\footnotesize
\begin{tabularx}{\linewidth}{@{}p{0.20\linewidth}YY@{}}
\toprule
Factor & CAD-guided measurement & CAD-unavailable or boundary-prior measurement \\
\midrule
Source coverage vs. source volume & Vary pose, lighting, material, clutter, and render count separately, then measure real-domain detection or pose transfer. & Vary normal-reference diversity, synthetic-anomaly assumptions, reference views, or feature-memory coverage separately. \\
Real calibration budget & Measure the marginal value of fixed real-label fractions for detector tuning, pose scoring, and acceptance thresholds. & Measure the marginal value of normal shots, few-shot references, or validation normals for threshold and mask calibration. \\
Verification channel & Compare appearance-only proposals with mask, depth, pose, and render-and-compare checks under the same proposals. & Compare feature, residual, semantic, depth, and multi-view evidence when no object-level model can be rendered. \\
Domain shift and threshold stability & Perturb lighting, camera pose, fixture, material, and occlusion while tracking geometric score calibration and latency. & Report false-alarm rate, missed-defect rate, mask calibration, and threshold transfer under controlled shifts. \\
\bottomrule
\end{tabularx}
\end{table}

\begin{figure}[H]
  \centering
  \includegraphics[width=\linewidth]{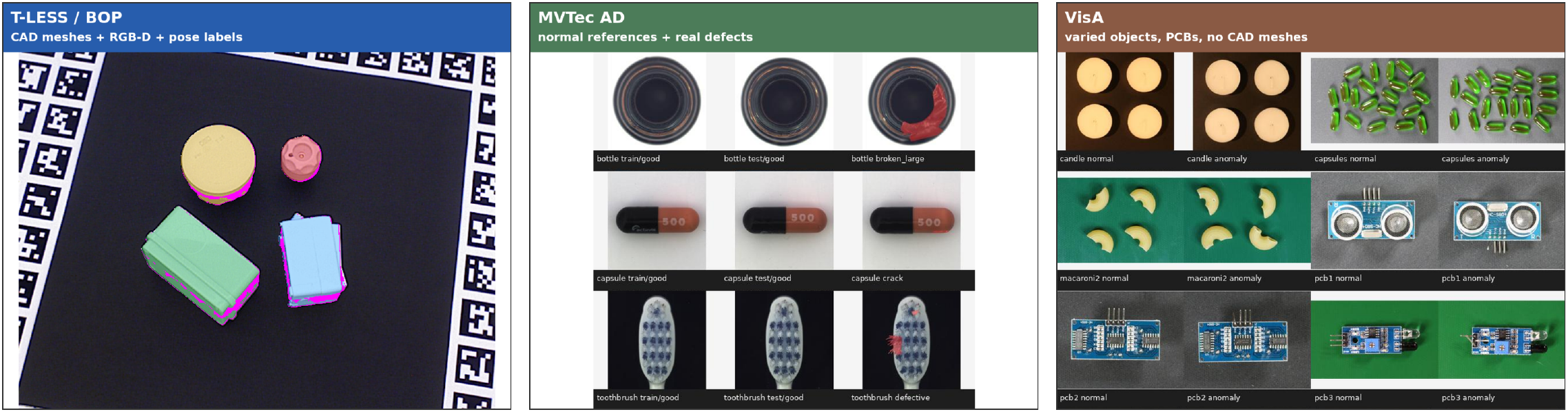}
  \caption{Dataset regimes used to anchor Section~\ref{sec:empirical-anchors}. T-LESS/BOP~\citep{hodan2017tless,hodan2018bop} represents the CAD-available branch, where object meshes, RGB-D observations, and pose labels make render-and-compare diagnostics possible. MVTec AD~\citep{bergmann2019mvtec} and VisA~\citep{zou2022visa} represent CAD-unavailable inspection, where the available evidence is normal-reference appearance and real anomalous test images rather than object-level geometry.}
  \label{fig:dataset-regimes}
\end{figure}

Figure~\ref{fig:dataset-regimes} provides the corresponding visual overview.
The T-LESS/BOP panel shows a real RGB observation with CAD rendered into the
camera view, making explicit why this branch can ask geometric verification
questions. The MVTec AD and VisA panels show the different CAD-unavailable evidence
structure: normal examples and real anomalous test cases are available, but
there is no object-level mesh to render back into the scene.

The visual evidence is organized in two layers. The main text shows the
mechanism needed for the review argument: selected validation predictions and
aggregate metrics for CAD-as-renderer transfer, CAD overlays and depth-fusion
examples for CAD-at-test-time verification, representative anomaly examples and
metric summaries for CAD-unavailable baselines, and the normal-reference budget trend.
The appendix keeps the heavier run-level diagnostics: YOLO training curves,
precision-recall curves, confusion matrices, validation prediction grids, CAD
overlay contact sheets, and synthetic-anomaly probe outputs. This prevents the
empirical section from becoming a benchmark report while preserving the
evidence trail behind the anchors.

\subsection{CAD-Guided Anchor: CAD as Synthetic Renderer}

The first CAD-available anchor treats CAD only as a source of synthetic RGB
labels. YOLOv8 detectors are trained on synthetic T-LESS PBR images and
evaluated on held-out real Primesense images \citep{ultralytics2023yolov8}.
In this block, object geometry is flattened into rendered training images and
is not used during inference. This isolates the renderer-only form of
CAD-available sim-to-real: the method can generate labeled source data, but it
cannot ask at test time whether a predicted box is geometrically consistent
with the object model.

This anchor is included because the CAD-as-renderer literature makes a
specific claim that should be separated from CAD-at-test-time geometry: if the
synthetic source distribution is well designed, CAD can reduce the need for
real labels before deployment. The anchor therefore varies the factors that the
literature identifies as central to renderer-based transfer. B0 and B1 ask
whether more PBR source images are sufficient by themselves. B2 represents
domain randomization as distributional coverage. B3 and B6 test whether a very
small labeled real set can calibrate residual appearance mismatch. B4 asks
whether weak real background evidence helps when object labels are still
synthetic. B5 asks whether detector capacity and pretrained representation
matter even when the source images are unchanged. YOLOv8 is used as a stable
industrial detector backbone for this controlled comparison, not because the
paper's claim depends on YOLO as the final object detector.

All detector runs share the same target domain: 900 held-out real T-LESS
Primesense images covering 30 object classes. The renderer-only block varies
the source-data recipe rather than the target evaluation set: synthetic source
size, train-time domain randomization, weak real background evidence, detector
capacity, and a small labeled real calibration set. The small-real-calibration
runs fine-tune on 50 labeled real Primesense images, corresponding to 5\% of
the real training split and 300 object annotations. This design makes the
block a transfer diagnostic rather than a search for a final detector recipe.

\begin{table}[H]
\centering
\caption{CAD-as-renderer detector baselines on real T-LESS images~\citep{hodan2017tless,hodan2018bop} using YOLOv8 detectors~\citep{ultralytics2023yolov8}. DR denotes domain randomization.}
\label{tab:cad-renderer}
\footnotesize
\begin{tabularx}{\linewidth}{@{}lp{0.19\linewidth}Xp{0.10\linewidth}lrr@{}}
\toprule
ID & Method & Training source & Real labels & Model & mAP$_{50}$ & mAP$_{50:95}$ \\
\midrule
B0 & Synthetic-only 5k & PBR 5k & No & YOLOv8n & 0.1987 & 0.1521 \\
B1 & Synthetic-only 50k & PBR 50k & No & YOLOv8n & 0.1667 & 0.1287 \\
B2 & Domain randomization & PBR 50k + DR augmentation & No & YOLOv8n & 0.5370 & 0.4041 \\
B3 & Small real fine-tune & B2 + 50 real images & Yes, 5\% & YOLOv8n & 0.7752 & 0.6265 \\
B4 & Real-background negatives & PBR 50k + DR + real backgrounds & Weak/no & YOLOv8n & 0.5410 & 0.4112 \\
B5 & Larger detector & PBR 50k + DR & No & YOLOv8s & 0.7762 & 0.6077 \\
B6 & Larger detector + real FT & B5 + 50 real images & Yes, 5\% & YOLOv8s & 0.8748 & 0.7424 \\
\bottomrule
\end{tabularx}
\end{table}

\begin{figure}[H]
  \centering
  \includegraphics[width=\linewidth]{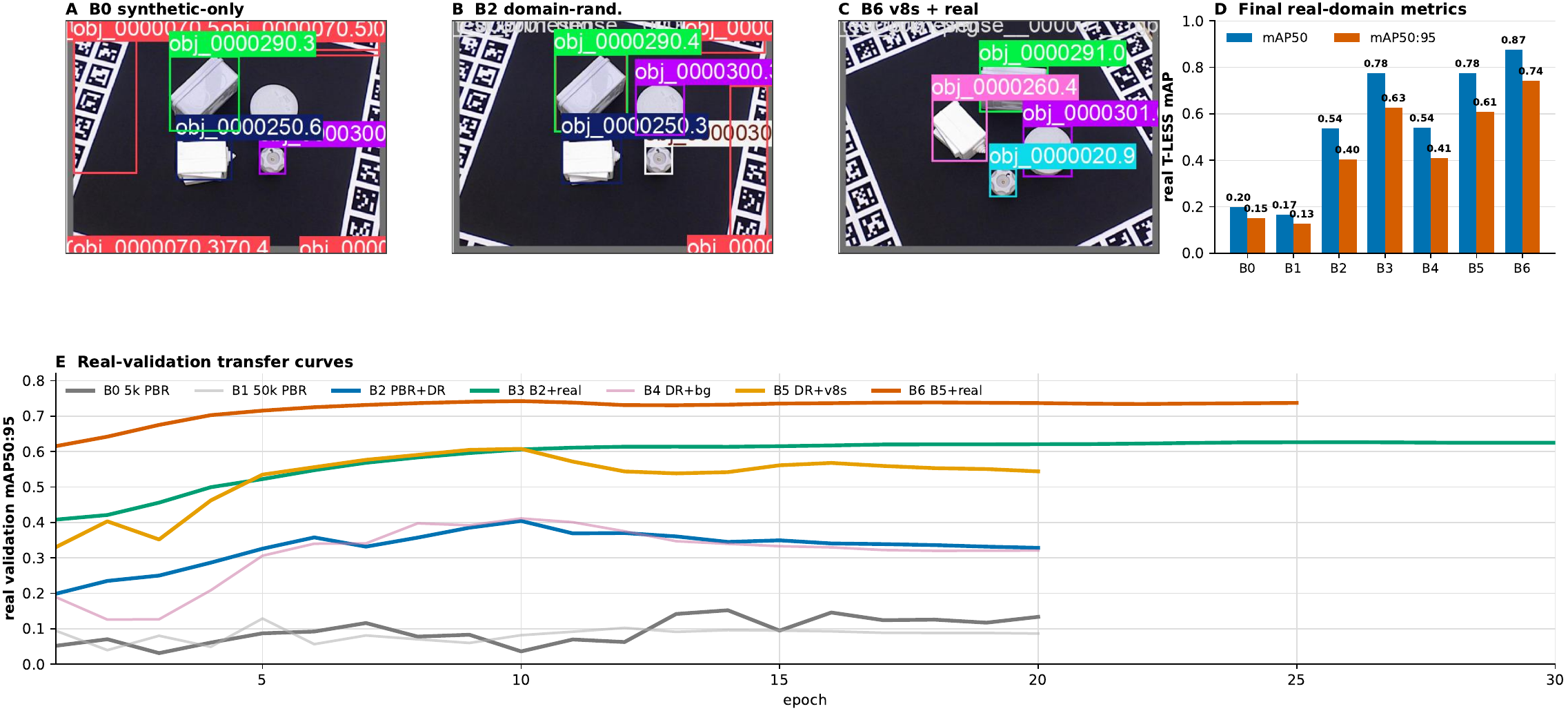}
  \caption{CAD-as-renderer transfer on real T-LESS images~\citep{hodan2017tless} using YOLOv8 detectors~\citep{ultralytics2023yolov8}. Panels A--C show representative validation predictions for synthetic-only training, domain-randomized training, and the strongest small-real-calibrated configuration. Panel D summarizes final B0--B6 detection metrics. Panel E shows real-validation mAP$_{50:95}$ across training, making clear that more synthetic images alone do not close transfer, whereas distribution design, detector capacity, and a small labeled real calibration set change real-domain behavior.}
  \label{fig:cad-renderer-transfer}
\end{figure}

\begin{figure}[H]
  \centering
  \includegraphics[width=\linewidth]{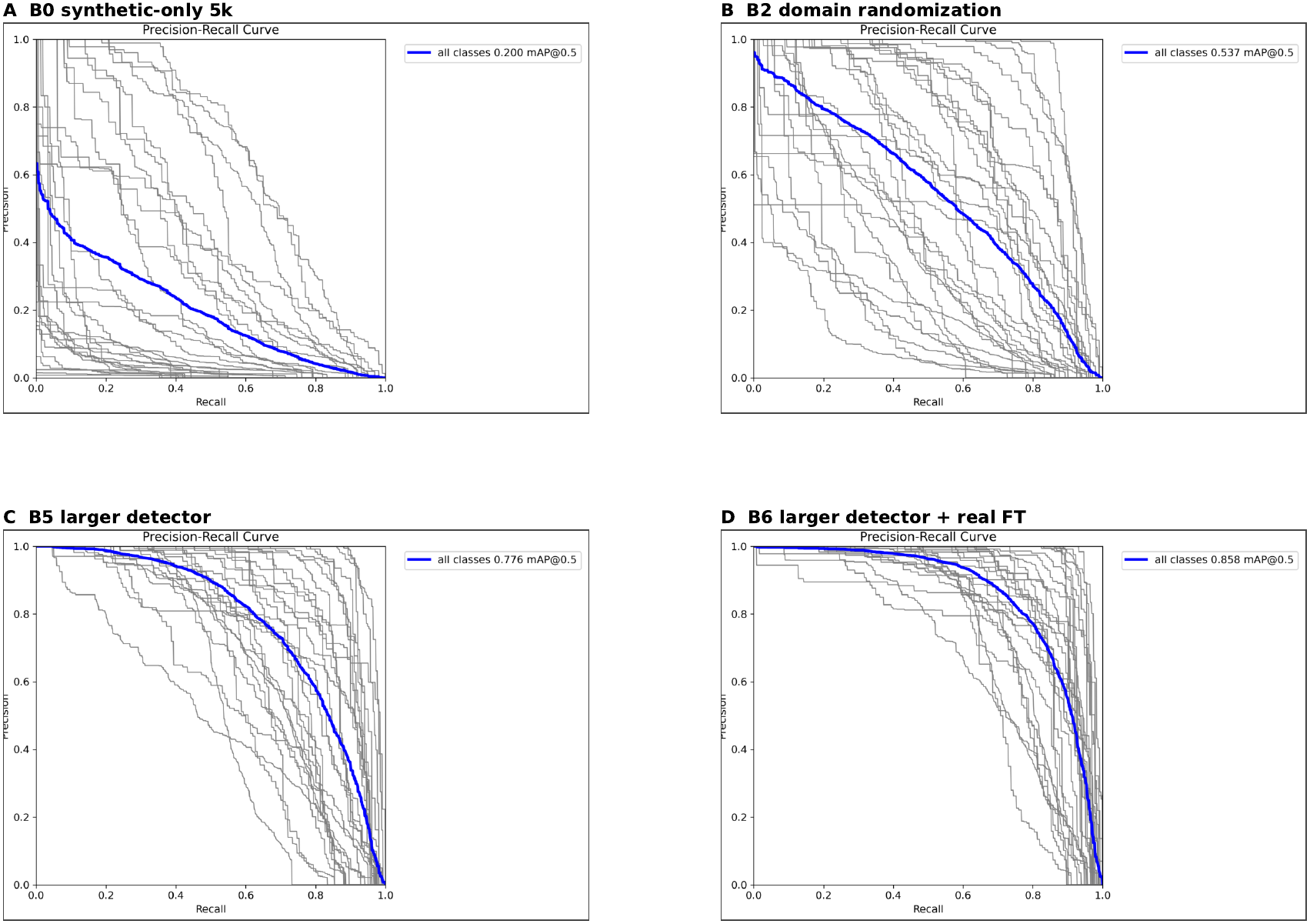}
  \caption{Precision-recall diagnostics for representative CAD-as-renderer detector runs on T-LESS~\citep{hodan2017tless} with YOLOv8~\citep{ultralytics2023yolov8}. B0 shows weak synthetic-only transfer, B2 shows the effect of domain randomization, B5 shows the larger-detector operating envelope under randomized synthetic training, and B6 shows the additional effect of small real fine-tuning. The blue curves report the all-class precision-recall summary, while the gray curves show class-specific variation.}
  \label{fig:cad-renderer-pr-curves}
\end{figure}

Figures~\ref{fig:cad-renderer-transfer} and
\ref{fig:cad-renderer-pr-curves} provide the visual counterpart to
Table~\ref{tab:cad-renderer}. Representative validation predictions show how
detector behavior changes from weak synthetic-only transfer to
domain-randomized transfer and then to the strongest small-real-calibrated
configuration. The real-validation curves show when each configuration
actually transfers during training, not only its final score, and the
precision-recall curves show how the operating envelope changes from
synthetic-only transfer to domain-randomized, larger-capacity, and
small-real-calibrated transfer. Confusion matrices, label distributions, and
full validation grids are retained in the appendix because they are useful
run-level diagnostics but too dense to carry the main argument alone.

Table~\ref{tab:cad-renderer} is best read as a transfer diagnostic. Increasing
synthetic PBR images from 5k to 50k does not improve real-domain accuracy;
mAP$_{50:95}$ decreases from 0.1521 to 0.1287. The first major improvement
comes from changing the source distribution: strong domain randomization raises
mAP$_{50:95}$ to 0.4041. This supports the review-level claim that CAD
rendering is a distribution-design problem rather than a raw image-count
problem.

Small real calibration and detector capacity then provide complementary ways
to absorb residual mismatch. Fine-tuning the domain-randomized YOLOv8n model
with only 50 labeled real images raises mAP$_{50:95}$ from 0.4041 to 0.6265.
Increasing detector capacity from YOLOv8n to YOLOv8s under synthetic
domain-randomized training reaches 0.6077 without real object labels.
Combining the larger detector with the same small real calibration set gives
0.7424 mAP$_{50:95}$. The review-level point is not that one detector variant
wins; it is that CAD-as-renderer transfer depends on the synthetic source
distribution, model capacity, and the role assigned to a small amount of real
evidence.

\subsection{CAD-Guided Anchor: CAD at Test Time}

The second CAD-available anchor asks what is gained when CAD remains available
during inference. A renderer-only detector uses geometry only before
deployment; a CAD-at-test-time system can additionally estimate pose, render
the object back into the camera, and score mask or depth consistency. MegaPose
is used as a practical render-and-compare representative
\citep{labbe2023megapose}. Newer CAD-aware foundation methods such as
FoundationPose, SAM-6D, GigaPose, FoundPose, and FreeZeV2 define the frontier
context, but the empirical role of MegaPose here is narrower: it instantiates a
runnable CAD-at-test-time pipeline rather than claiming to represent the latest
state of the art
\citep{wen2024foundationpose,lin2024sam6d,nguyen2024gigapose,foundpose2024,freezev2}.

This anchor follows directly from the CAD-available literature reviewed above:
CAD is valuable not only because it can make synthetic labels, but because it
can make real test-time hypotheses geometrically checkable. A detector box is
an appearance hypothesis; a CAD pose and rendered mask turn it into a geometric
hypothesis. MegaPose is therefore used to instantiate the render-and-compare
mechanism, while the oracle and bridge diagnostics separate three questions
that would otherwise be conflated: whether the dataset geometry is internally
consistent, whether the detector supplies usable proposals for a CAD pipeline,
and whether practical estimated-pose render consistency can approximate the
ideal value of geometry.

The CAD-at-test-time diagnostics are built on the same held-out real T-LESS
split but ask a different question from the renderer-only detector experiment.
First, an oracle geometry check renders each T-LESS CAD model from the BOP
ground-truth pose and camera intrinsics, then compares the rendered CAD mask
with the BOP full and visible masks. This tests whether the meshes, camera
metadata, pose labels, and renderer are consistent enough for render-and-compare
analysis. Second, the B6 detector is used as the practical ROI source, and its
detections are evaluated as a bridge into CAD inference through same-class ROI
recall. Third, an oracle CAD-verification score is computed for B6 proposals to
measure the upper-bound value of an ideal geometric consistency signal.

\begin{figure}[H]
  \centering
  \includegraphics[width=\linewidth]{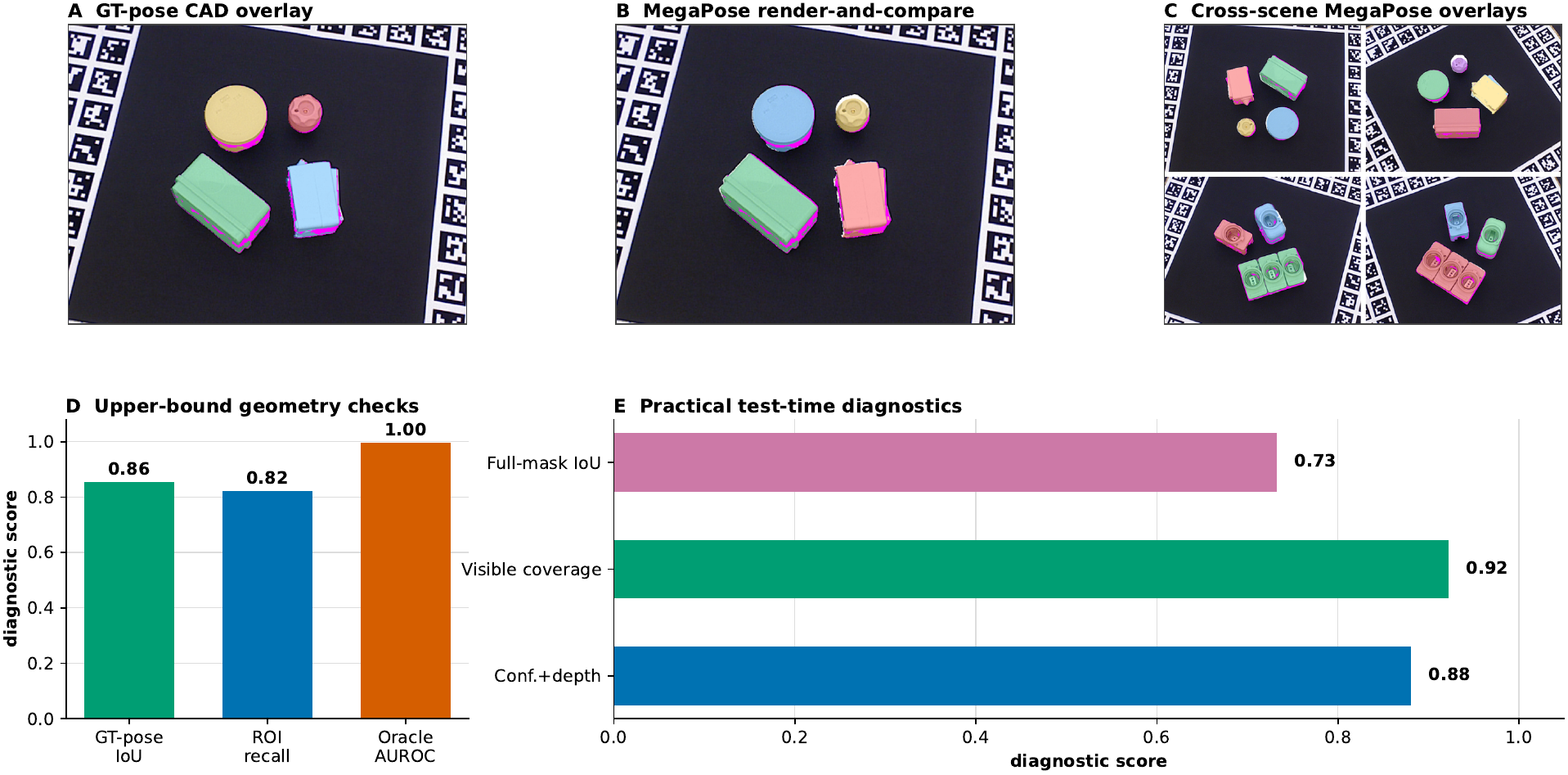}
  \caption{CAD-at-test-time geometry on T-LESS/BOP~\citep{hodan2017tless,hodan2018bop}. Panels A and B show CAD render overlays from ground-truth pose and MegaPose-style render-and-compare inference~\citep{labbe2023megapose}. Panel C shows additional cross-scene MegaPose overlays, illustrating that the CAD prior is repeatedly rendered back into real camera views rather than used only as a training source. Panel D summarizes upper-bound geometry diagnostics, and Panel E summarizes practical test-time diagnostics. Together, the panels show how CAD turns part of the domain gap into a geometric consistency check, while practical performance still depends on proposal quality and pose estimation.}
  \label{fig:cad-test-time-geometry}
\end{figure}

The practical render-and-compare run then uses B6 boxes, predicted class
labels, camera intrinsics, and the corresponding T-LESS CAD meshes as MegaPose
inputs. The main batch contains 100 real Primesense images sampled across 20
scenes, with up to 10 detector proposals per image, giving 689 proposals after
filtering. MegaPose estimates 6D pose for each proposal; the predicted pose is
rendered back into the image to compute full-mask IoU, scene-visible-mask IoU,
and ground-truth visible coverage. A final diagnostic compares real T-LESS
depth with CAD depth rendered from the MegaPose pose and fuses that depth
consistency score with detector confidence. Thus
Figure~\ref{fig:cad-test-time-geometry} summarizes both the visual output of
the render-and-compare procedure and the numeric diagnostics that result from
the pose, mask, and depth pipeline.

The metrics are chosen to match this staged question. Visible-mask IoU from
ground-truth pose tests the upper-bound consistency of CAD models, camera
metadata, and masks. ROI recall tests whether a detector can feed a downstream
CAD-at-test-time method; missed proposals cannot be recovered by a pose
estimator. Proposal AUROC asks whether a score can separate good and bad
hypotheses, which is the practical purpose of verification. Full-mask IoU and
visible coverage summarize how well the estimated CAD pose explains the real
object region. Depth-fusion AUROC then tests whether metric geometry adds
information beyond detector confidence. A full BOP-style pose benchmark could
also report VSD, MSSD, MSPD, ADD(-S), or recall under pose-error thresholds,
but those metrics would answer a pose-leaderboard question. Here the narrower
question is whether CAD at test time provides a verification channel that
renderer-only detection lacks.

\begin{table}[H]
\centering
\caption{CAD-at-test-time diagnostics on T-LESS/BOP~\citep{hodan2017tless,hodan2018bop} using oracle CAD rendering, YOLOv8 detector proposals~\citep{ultralytics2023yolov8}, and MegaPose render-and-compare inference~\citep{labbe2023megapose}.}
\label{tab:cad-test-time}
\small
\begin{tabularx}{\linewidth}{lXlr}
\toprule
Diagnostic & Scale & Metric & Result \\
\midrule
GT-pose CAD oracle & 900 images / 6,169 instances & mean visible-mask IoU & 0.8553 \\
B6 ROI bridge & 900 images / confidence 0.10 & same-class ROI recall@0.5 & 0.8206 \\
Oracle CAD verification & 8,042 proposals / confidence 0.10 & proposal AUROC & 0.9963 \\
MegaPose cross-scene & 100 images / 689 proposals & mean full-mask IoU & 0.7322 \\
MegaPose same-class nonzero & 573 proposals & mean GT-visible coverage & 0.9218 \\
Confidence + depth fusion & 100 images / 689 proposals & good-pose AUROC & 0.8804 \\
\bottomrule
\end{tabularx}
\end{table}

Figure~\ref{fig:cad-test-time-geometry} visualizes the second role of CAD:
geometry can be rendered back into the real camera view and scored, rather than
only used to make synthetic training images. This is the main reason the
CAD-available branch is not reducible to synthetic-data generation.

Table~\ref{tab:cad-test-time} separates upper-bound geometry from practical
geometry. The ground-truth pose oracle establishes that the T-LESS CAD models,
camera metadata, and masks are internally reliable: rendering CAD models from
BOP ground-truth poses yields 0.8553 mean visible-mask IoU. The detector-to-CAD
bridge then shows where a deployed pipeline begins to lose this oracle
structure: the best detector supplies usable but imperfect regions of
interest, reaching same-class ROI recall@0.5 of 0.8206 at confidence 0.10.
Oracle CAD verification shows why geometry is valuable in principle, because
an ideal CAD score separates true and false proposals with AUROC 0.9963.

The practical MegaPose and depth-fusion results are deliberately interpreted
more cautiously. On a 100-image cross-scene batch, MegaPose reaches 0.7322
mean full-mask IoU across all proposals, including detector false positives
and hard object cases. Among same-class nonzero-alignment proposals, mean
ground-truth visible coverage reaches 0.9218, showing that practical CAD pose
can strongly cover visible object regions when the detector supplies a valid
proposal. Rendered-depth consistency is not a standalone replacement for
detector confidence, but fusing depth consistency with confidence improves
good-pose AUROC to 0.8804. Thus, CAD at test time adds a verification channel
that renderer-only detection lacks, while still inheriting proposal errors,
pose failures, occlusion, and object ambiguity from the real deployment scene.

\subsection{CAD-Unavailable Anchor: Normality and Foundation Priors}

The CAD-unavailable branch switches from known-object detection and pose
diagnostics to industrial anomaly detection. Here the system cannot render the
target object, estimate object pose from a mesh, or verify a hypothesis against
CAD geometry. The empirical question is therefore not how well a detector
transfers from rendered CAD labels, but which non-geometric prior can replace
the missing model. All main CAD-unavailable baselines use normal training images only
and do not use real anomaly labels for training.

The method choices follow the CAD-unavailable literature families reviewed earlier
rather than a single leaderboard logic. PatchCore represents normal-reference
memory, the most direct substitute for geometry because it stores what normal
real appearance looks like \citep{roth2022patchcore}. EfficientAD-S represents
compact teacher-student residual modeling, a different industrial prior that
emphasizes dense anomaly signals and fast inference \citep{batzner2024efficientad}.
WinCLIP represents the semantic zero-shot question: whether language-aligned
vision priors can reduce object-specific training requirements
\citep{jeong2023winclip}. AnomalyDINO represents a stronger spatial foundation
feature prior built from dense DINOv2 features
\citep{oquab2023dinov2,damm2025anomalydino}. DRAEM and
SuperSimpleNet-style synthetic-anomaly runs are kept as probes of
synthetic-anomaly self-supervision, but they are not promoted to full
all-category anchors because the completed probes were less stable than the
normal-reference and foundation-feature families
\citep{zavrtanik2021draem,liu2023simplenet,rolih2024supersimplenet}.

The CAD-unavailable anchors are run category-wise on the standard MVTec AD and VisA
splits, then macro-averaged over 15 MVTec AD categories and 12 VisA categories.
The implementation settings are fixed to keep the comparison focused on prior
families: PatchCore is fit as a normal-memory method; EfficientAD-S is trained
as a compact teacher-student model; WinCLIP is evaluated in zero-shot mode
($k=0$); and AnomalyDINO uses DINOv2 dense patch features as a normal-image
memory bank. The primary AnomalyDINO run uses
\texttt{dinov2\_vit\_small\_14}, while the scale ablation uses
\texttt{dinov2\_vit\_large\_14}. These settings are not an exhaustive
hyperparameter search. They are chosen to compare the kinds of priors that can
stand in for geometry when CAD is absent.

The metric set is also chosen to keep the CAD-unavailable comparison tied to inspection
rather than to a single benchmark convention. Image AUROC asks whether abnormal
products are ranked above normal products independent of a chosen threshold.
Pixel AUROC asks whether the anomaly map ranks defective pixels above normal
pixels, which is closer to localization but remains threshold-free. Image F1
and pixel F1 expose the separate operating-point question: whether those scores
can be converted into usable binary decisions and masks. A deployment study
could additionally use PRO, average precision, false positives per image,
missed-defect rate at a fixed false-alarm budget, calibration error, or latency.
Those metrics are important, but they require deployment cost assumptions that
are outside this review anchor. AUROC/F1 are used here because they are common
across MVTec AD and VisA and separate ranking quality from thresholded behavior.
For deployment, the next layer would be threshold stability: whether a
threshold chosen on normal validation data preserves a specified false-alarm
budget after lighting, material, fixture, or category shifts. Mask calibration
should likewise be reported as a decision property, not only as pixel ranking:
the relevant question is whether highlighted regions remain useful for reject,
repair, or human-review decisions under the expected operating cost.

\begin{table}[H]
\centering
\caption{CAD-unavailable anomaly detection anchors on MVTec AD~\citep{bergmann2019mvtec} and VisA~\citep{zou2022visa}. Methods instantiate normal memory (PatchCore~\citep{roth2022patchcore}), teacher-student residuals (EfficientAD~\citep{batzner2024efficientad}), zero-shot vision-language transfer (WinCLIP~\citep{jeong2023winclip}), and dense foundation features (DINOv2/AnomalyDINO~\citep{oquab2023dinov2,damm2025anomalydino}). AUROC reports ranking behavior; F1 reports thresholded decision and mask behavior.}
\label{tab:no-cad-baselines}
\footnotesize
\begin{tabularx}{\linewidth}{@{}p{0.15\linewidth}p{0.18\linewidth}p{0.10\linewidth}rrrr@{}}
\toprule
Method & Family & Dataset & Img. AUROC & Img. F1 & Pix. AUROC & Pix. F1 \\
\midrule
PatchCore & Normal memory & MVTec AD & 0.9820 & 0.9711 & 0.9801 & 0.5806 \\
PatchCore & Normal memory & VisA & 0.9080 & 0.8704 & 0.9807 & 0.4255 \\
EfficientAD-S & Teacher-student & MVTec AD & 0.9749 & 0.9553 & 0.9512 & 0.6265 \\
EfficientAD-S & Teacher-student & VisA & 0.9153 & 0.8838 & 0.9766 & 0.4057 \\
WinCLIP & Zero-shot VLM & MVTec AD & 0.8812 & 0.9185 & 0.6204 & 0.0891 \\
WinCLIP & Zero-shot VLM & VisA & 0.7553 & 0.7779 & 0.5900 & 0.0194 \\
AnomalyDINO-S & Dense foundation & MVTec AD & 0.9793 & 0.9652 & 0.9700 & 0.5707 \\
AnomalyDINO-S & Dense foundation & VisA & 0.9109 & 0.8654 & 0.9568 & 0.4199 \\
AnomalyDINO-L & Scale ablation & MVTec AD & 0.9814 & 0.9670 & 0.9596 & 0.5281 \\
AnomalyDINO-L & Scale ablation & VisA & 0.9330 & 0.8869 & 0.9690 & 0.4076 \\
\bottomrule
\end{tabularx}
\end{table}

\begin{figure}[H]
  \centering
  \includegraphics[width=\linewidth]{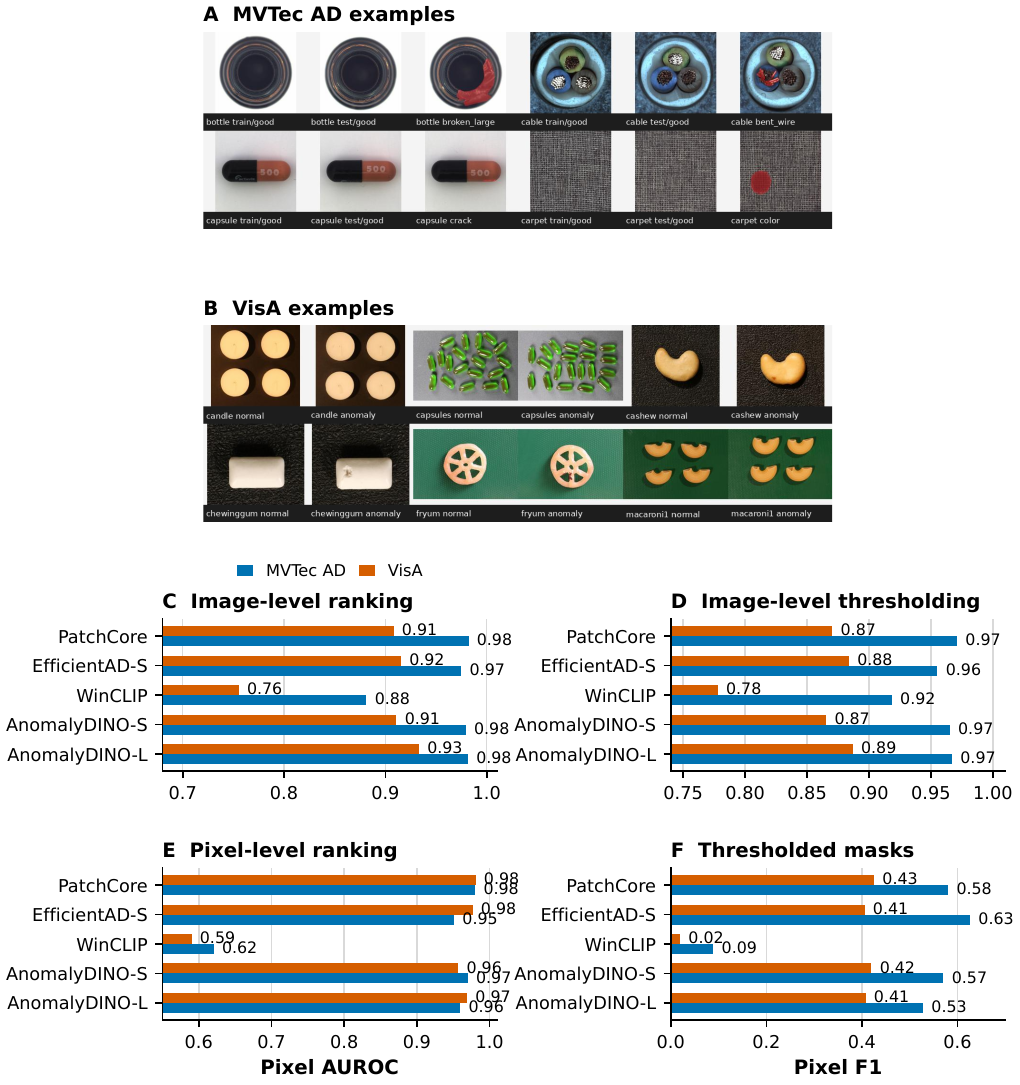}
  \caption{CAD-unavailable anomaly-detection anchors on MVTec AD~\citep{bergmann2019mvtec} and VisA~\citep{zou2022visa}. Panels A and B show representative normal/anomalous examples. Panels C--F summarize image AUROC, image F1, pixel AUROC, and pixel F1 for PatchCore~\citep{roth2022patchcore}, EfficientAD~\citep{batzner2024efficientad}, WinCLIP~\citep{jeong2023winclip}, and AnomalyDINO~\citep{damm2025anomalydino}, keeping product-level ranking, dense localization, and thresholded behavior visible in one view.}
  \label{fig:no-cad-anomaly-anchors}
\end{figure}

Figure~\ref{fig:no-cad-anomaly-anchors} places the benchmark numbers next to
representative normal and anomalous examples so that the reader sees what kind
of evidence the CAD-unavailable methods receive. The four metric panels are included
not to multiply scores, but to make one review-level point visible: product
ranking, dense localization, and thresholded mask quality are different
questions when geometry is unavailable.

Table~\ref{tab:no-cad-baselines} and
Figure~\ref{fig:no-cad-anomaly-anchors} are therefore best read by prior type.
Normal-reference methods remain strong because they turn real normal images
into the substitute for explicit object geometry: PatchCore gives the best
MVTec AD image AUROC and the best pixel AUROC on both datasets, while
EfficientAD-S gives a different operating point with stronger MVTec AD pixel F1
and slightly stronger VisA image AUROC. Zero-shot language-aligned transfer is
much weaker for dense industrial localization under this protocol. WinCLIP
retains moderate image-level signal on MVTec AD, but its pixel scores fall far
below the normal-reference methods, especially on VisA. Dense visual foundation
features are more competitive: AnomalyDINO-S approaches PatchCore and
EfficientAD-S on MVTec AD, and the DINOv2-Large ablation gives the strongest
VisA image AUROC. The larger backbone does not, however, monotonically improve
pixel localization or thresholded masks, so the foundation-feature conclusion
is about useful spatial priors rather than scale alone.

\begin{figure}[H]
  \centering
  \includegraphics[width=\linewidth]{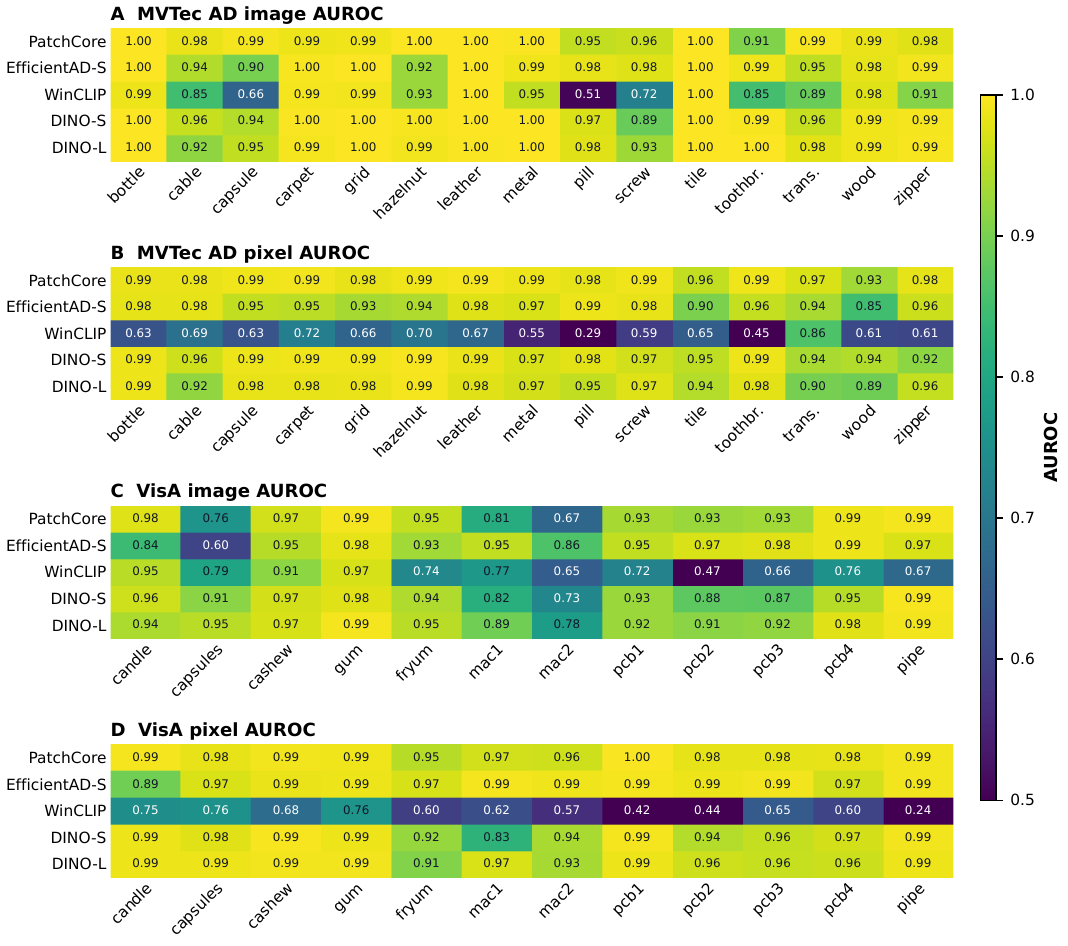}
  \caption{Category-level CAD-unavailable diagnostics for MVTec AD~\citep{bergmann2019mvtec} and VisA~\citep{zou2022visa}. Each cell reports AUROC for one method and category. Panels A and B show MVTec AD image and pixel AUROC; panels C and D show VisA image and pixel AUROC. The category view shows where the macro averages come from and highlights the persistent pixel-localization weakness of WinCLIP-style zero-shot transfer~\citep{jeong2023winclip}.}
  \label{fig:no-cad-category-diagnostics}
\end{figure}

Figure~\ref{fig:no-cad-category-diagnostics} adds the category-level view
needed for a review paper. CAD-unavailable inspection is sensitive to whether the
category is a repeated product, a texture-like surface, a small-defect object,
or a PCB-style layout. The heat maps show that the macro means are not driven
by a single easy category family, and they make the zero-shot VLM limitation
visible across categories at the pixel level rather than only as a low average.

\subsection{CAD-Unavailable Normal-Reference Budget Diagnostic}

The previous CAD-unavailable anchor compares prior families under full normal-reference
access. A second, narrower diagnostic asks a more practical question: when CAD
is absent, how quickly does a small set of normal real images become useful?
This is the CAD-unavailable analogue of the small-real-calibration question in the
CAD-as-renderer branch, but the role of real data is different. The images do
not calibrate a rendered source domain; they are the reference evidence from
which normality is defined.

The budget diagnostic therefore subsamples only the normal training split; the
test images, anomaly labels, and pixel masks are unchanged. PatchCore and
AnomalyDINO-S are evaluated on three MVTec AD categories and three VisA
categories using 5\%, 10\%, 25\%, and 100\% of normal training images. The
selected MVTec AD categories are toothbrush, capsule, and cable, and the
selected VisA categories are macaroni2, capsules, and pcb1. These categories
cover a low-count object case, structured objects, a hard VisA category, and a
PCB inspection case. The 5\%, 10\%, and 25\% rows are new budgeted fits, while
the 100\% rows reuse the corresponding full-baseline results to keep the
diagnostic aligned with Table~\ref{tab:no-cad-baselines}.

\begin{table}[H]
\centering
\caption{Selected-category CAD-unavailable normal-reference budget diagnostic on MVTec AD~\citep{bergmann2019mvtec} and VisA~\citep{zou2022visa}. PatchCore~\citep{roth2022patchcore} and AnomalyDINO~\citep{damm2025anomalydino} are evaluated with reduced normal-reference sets. Image and pixel columns report AUROC.}
\label{tab:no-cad-budget}
\small
\begin{tabular}{llcccc}
\toprule
 & & \multicolumn{2}{c}{5\% normal} & \multicolumn{2}{c}{100\% normal} \\
\cmidrule(lr){3-4}\cmidrule(lr){5-6}
Method & Dataset & Image & Pixel & Image & Pixel \\
\midrule
PatchCore & MVTec AD & 0.9020 & 0.9819 & 0.9613 & 0.9879 \\
PatchCore & VisA & 0.7509 & 0.9712 & 0.7872 & 0.9813 \\
AnomalyDINO-S & MVTec AD & 0.9445 & 0.9732 & 0.9655 & 0.9800 \\
AnomalyDINO-S & VisA & 0.8306 & 0.9642 & 0.8573 & 0.9693 \\
\bottomrule
\end{tabular}
\end{table}

\begin{figure}[H]
  \centering
  \includegraphics[width=\linewidth]{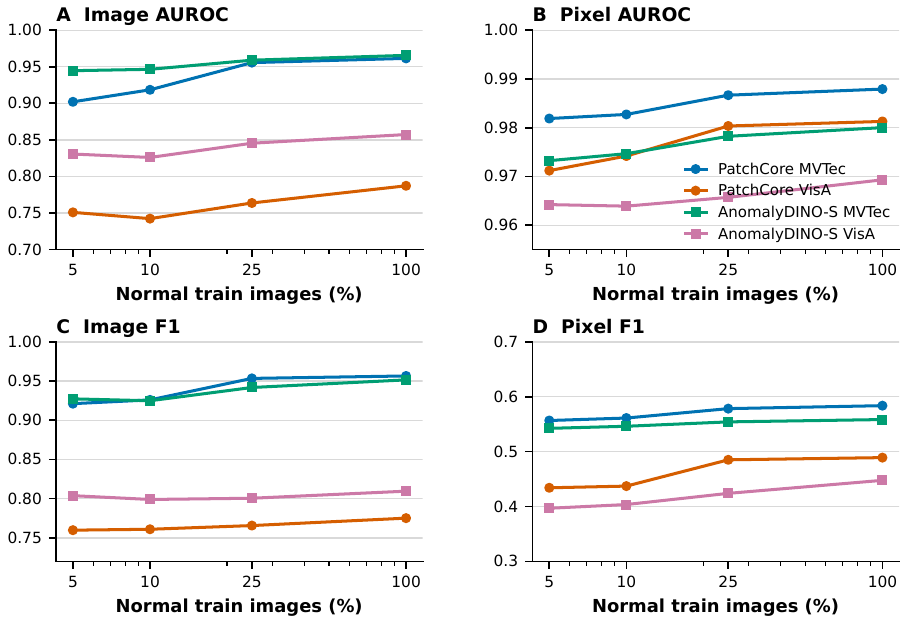}
  \caption{Selected-category normal-reference budget curves for the CAD-unavailable branch. PatchCore~\citep{roth2022patchcore} and AnomalyDINO-S~\citep{damm2025anomalydino} are evaluated with 5\%, 10\%, 25\%, and 100\% of the normal training images on selected MVTec AD~\citep{bergmann2019mvtec} and VisA~\citep{zou2022visa} categories. Pixel AUROC remains high even with small normal sets, while image-level discrimination and thresholded F1 improve more gradually with reference coverage.}
  \label{fig:no-cad-budget-curves}
\end{figure}

Figure~\ref{fig:no-cad-budget-curves} shows the resulting sensitivity to normal
reference coverage. Because the test set is fixed, the budget axis is not a new
difficulty setting; it changes only how much normal appearance evidence the
method can store or compare against.

Table~\ref{tab:no-cad-budget} and
Figure~\ref{fig:no-cad-budget-curves} suggest that CAD-unavailable methods can be
surprisingly data-efficient for pixel AUROC. Even with 5\% of normal training
images, PatchCore reaches 0.9819 pixel AUROC on selected MVTec AD categories
and 0.9712 on selected VisA categories. AnomalyDINO-S similarly remains strong
at 5\%, especially in image AUROC on VisA. Larger normal-reference budgets
mainly improve image-level discrimination and thresholded masks. The diagnostic
therefore connects back to the review taxonomy: real data help both branches,
but in CAD-available detection they calibrate transfer from rendered source
images, whereas in CAD-unavailable anomaly detection they become the reference model
itself.

\subsection{Cross-Regime Synthesis}

The empirical anchors now fold back into the review structure. Industrial
sim-to-real performance is governed less by ``synthetic versus real'' as a
binary distinction than by the prior available for transfer, calibration, and
verification. In the CAD-available branch, more synthetic images alone do not
solve transfer; domain randomization, model capacity, and small real
calibration matter more. When CAD remains available at test time, pose,
rendering, and depth consistency provide an additional way to check real
observations. In the CAD-unavailable branch, normal-reference memory and dense
visual foundation features provide the most reliable appearance-only
alternatives, while zero-shot language-aligned priors and simple
synthetic-anomaly probes are weaker for dense industrial defects under the
present protocols.

Table~\ref{tab:cross-regime-synthesis} is therefore not an additional results
table. It summarizes how the anchors support the review's organizing claim
that industrial sim-to-real strategies depend on the prior available at
deployment.

\begin{table}[H]
\centering
\caption{Interpretive synthesis of the two empirical branches.}
\label{tab:cross-regime-synthesis}
\small
\begin{tabularx}{\linewidth}{p{0.22\linewidth}XX}
\toprule
Question & CAD available & CAD unavailable \\
\midrule
Does generated data alone solve transfer? & PBR 50k does not improve over PBR 5k; distribution design matters. & Generated-anomaly probes are useful diagnostics, but they do not replace stronger normal-reference and foundation-feature anchors here. \\
What improves real-domain robustness? & Domain randomization, detector capacity, and 5\% real fine-tuning stack strongly. & Normal-reference memory and dense DINOv2 features are strongest; zero-shot CLIP-style transfer is weak for pixels. \\
How does small real data help? & Small labeled real data calibrates supervised synthetic-to-real detection. & Small normal-reference sets define an appearance memory and preserve high pixel AUROC. \\
What does geometry add? & CAD pose, rendering, and depth consistency provide test-time verification. & CAD-unavailable methods cannot render object geometry and must rely on feature deviation or normality. \\
What remains hard? & Proposal errors, pose failures, object ambiguity, occlusion, and practical scoring. & VisA image-level difficulty, threshold calibration, dense masks, and deployment shifts. \\
\bottomrule
\end{tabularx}
\end{table}

The empirical evidence therefore supports the two-situation review structure.
CAD availability should not be treated as a generic advantage; its value
depends on whether it is used as a renderer, a calibration scaffold, or a
test-time geometric prior. CAD-unavailable inspection should not be treated as
merely weaker or less informed; it uses a different evidence source, namely
normal-reference appearance and pretrained feature spaces. The experiments make
these differences visible while deliberately avoiding a direct numerical
comparison between T-LESS detection metrics and MVTec/VisA anomaly metrics.
Boundary-prior methods should be read through the same lens: the question is
whether their partial prior supports rendering, weak matching, semantic
localization, or calibration, not whether they form a single benchmarkable
third category.

\section{Conclusion and Outlook}

Industrial visual sim-to-real is often introduced as a problem of synthetic
images transferring to real images. This review argues for a broader and more
useful formulation. In industrial vision, the relevant gap is between the
evidence available before or during deployment and the evidence required to
make reliable decisions in the real production environment. CAD availability
is therefore not a minor dataset attribute. It determines whether
the system can use explicit geometry as a source of labels, as a calibration
scaffold, or as a test-time verification signal.

When CAD or a renderable object model is available, sim-to-real should be
understood as a geometry-aware transfer problem. CAD can generate labeled
source data, but the empirical anchors show that render count alone is not the
main issue. Source-distribution design, detector capacity, and small amounts
of real calibration can matter more than simply increasing synthetic volume.
If CAD remains available at test time, the problem changes again: pose,
rendering, mask overlap, and depth consistency can turn an appearance
hypothesis into a geometrically checkable hypothesis. This does not remove
proposal errors, pose ambiguity, occlusion, or scoring difficulty, but it
provides a verification channel that renderer-only training cannot provide.

When CAD is unavailable, the same industrial deployment pressure leads to a
different class of priors. CAD-unavailable inspection methods must decide whether an
observation is consistent with normal appearance, feature distributions,
teacher-student residuals, synthetic-anomaly assumptions, semantic prompts, or
dense foundation features. The empirical anchors reinforce the literature
review: normal-reference memory and dense visual foundation features are strong
appearance-only substitutes for explicit geometry, while zero-shot
language-aligned transfer remains less reliable for dense localization under
standard anomaly protocols. Small normal-reference sets can already provide
useful pixel-level ranking, but thresholded masks, image-level decisions, and
category-specific behavior remain central deployment challenges.

The boundary between the two situations will continue to blur. Approximate
models, reference-light pose methods, CAD-template segmentation, dense
foundation features, generative anomaly synthesis, and large vision-language
systems all weaken strict assumptions about what is known before deployment.
These directions form a boundary-prior regime: they provide intermediate
forms of evidence rather than full CAD geometry or complete CAD absence. They
do not, however, remove the organizing question of the review: what prior
grounds the decision? A generated defect, a prompt, a reference image, a
learned feature space, and a CAD mesh provide different kinds of support, and
they should not be evaluated as if they were interchangeable.

The practical implication is that future industrial sim-to-real evaluations
should state the available prior before selecting methods, datasets, or
metrics. If CAD is available only before deployment, evaluation should focus on
source-distribution design and real-domain calibration. If CAD remains
available at test time, evaluation should include geometric verification
channels. If CAD is absent, evaluation should report how normal-reference
coverage, foundation features, threshold calibration, and category structure
affect inspection behavior. A review organized around prior availability does
not replace task-specific benchmarks; it explains why those benchmarks answer
different questions. This is the basis for comparing industrial sim-to-real
methods without collapsing detection, pose, and anomaly inspection into a
single cross-regime scoreboard.

Test-time adaptation, self-training, and uncertainty estimation fit naturally
into this same view. They do not define a separate prior regime; rather, they
change how a method uses deployment observations after the initial prior has
been specified. In CAD-guided systems, adaptation can tune proposal scores,
appearance features, or render-and-compare thresholds, but it should remain
constrained by geometric consistency so that adaptation does not reinforce
incorrect poses. In CAD-unavailable inspection, adaptation can update normal
feature memories or thresholds, but it must be monitored for defect leakage and
threshold drift. Uncertainty estimates are most useful when tied to an action:
accept, reject, request another view, lower line speed, or ask for human
review.

\begin{table}[H]
\centering
\caption{Practitioner reporting checklist for industrial visual sim-to-real studies.}
\label{tab:practitioner-checklist}
\footnotesize
\begin{tabularx}{\linewidth}{@{}p{0.24\linewidth}Y@{}}
\toprule
Item to report & Purpose \\
\midrule
Available prior & State whether the method uses full CAD, approximate geometry, templates, reference views, normal images, synthetic anomalies, pretrained features, prompts, or combinations of these. \\
Evidence channel & Specify whether the prior supports source generation, correspondence, test-time checking, calibration, or only semantic/appearance evidence. \\
Calibration budget & Report the number and type of real labels, normal references, validation normals, or threshold-tuning examples used. \\
Operating point & Report the thresholding rule, expected false-alarm budget, missed-defect tolerance, and whether the same threshold transfers across shifts. \\
Stress audit & Identify failures under occlusion, symmetries, transparent or reflective materials, clutter, multi-view disagreement, latency limits, and production shifts. \\
\bottomrule
\end{tabularx}
\end{table}

The central recommendation is therefore simple: report industrial sim-to-real
results with the prior, the evidence channel, and the deployment operating
point made explicit. Doing so turns prior availability from a descriptive
taxonomy into a practical design and evaluation principle.

\section*{Code and Data Availability}

The accompanying project repository provides the review support materials,
metadata, and reproducibility artifacts used to organize the empirical anchors:
\url{https://github.com/JacksonTao888/industrial-visual-sim2real-priors}.

\clearpage
\bibliographystyle{unsrtnat}
\bibliography{references}

\clearpage
\appendix
\section{Empirical Diagnostics for the Review Anchors}
\label{app:empirical-diagnostics}

This appendix preserves run-level diagnostics behind the empirical anchors in
Section~\ref{sec:empirical-anchors}. Its purpose is reproducibility and audit
support: the figures document the completed detector, CAD-at-test-time, and
CAD-unavailable probe outputs that underlie the main-text tables and figures.
They do not introduce a new benchmark, a new method family, or a separate
comparison beyond the review anchors. The appendix is therefore best read as
supporting evidence for the paper's prior-availability framing rather than as
a standalone experimental section.

Epoch-level training curves are available for the YOLOv8 detector block~\citep{ultralytics2023yolov8} because
those runs share a supervised training protocol. The CAD-at-test-time and
CAD-unavailable anchors do not all produce comparable epoch curves: pose and
CAD-consistency runs are inference-time geometry checks, while several CAD-unavailable
inspection baselines are memory, feature, or probe-style workflows. For those
anchors, the appendix therefore preserves qualitative contact sheets rather
than forcing unlike methods into the same training-curve format.

\clearpage
\begin{figure}[H]
\centering
\includegraphics[width=\linewidth]{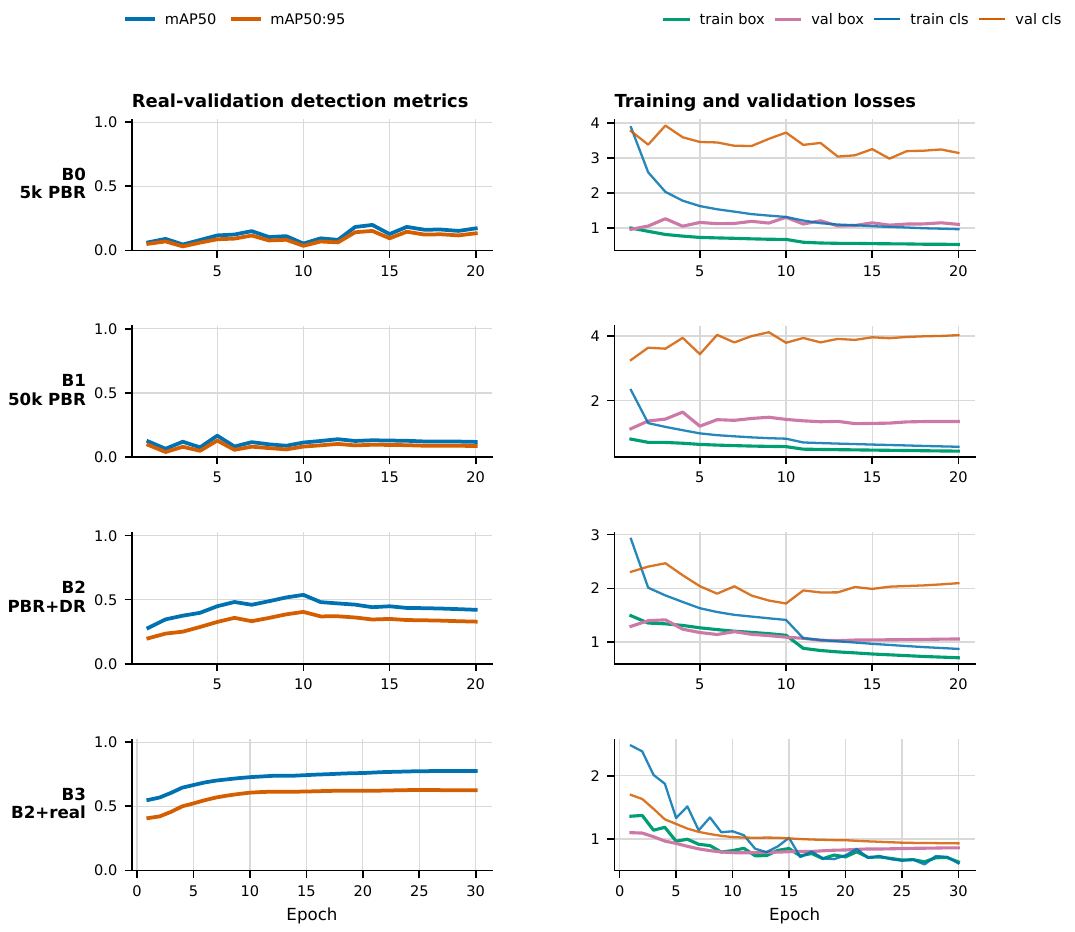}
\caption{YOLOv8~\citep{ultralytics2023yolov8} CAD-as-renderer training diagnostics for B0--B3 on T-LESS/BOP~\citep{hodan2017tless,hodan2018bop}. Each row
corresponds to one detector configuration from
Table~\ref{tab:cad-renderer}. The left column reports real-domain
validation mAP during training; the right column reports training and
validation box/classification losses. B0 and B1 show that larger PBR-only
synthetic volume does not by itself close the real-domain gap, B2 shows the
effect of domain randomization, and B3 shows the calibration effect of a small
real fine-tuning set.}
\label{fig:app-yolo-training-b0-b3}
\end{figure}

\begin{figure}[H]
\centering
\includegraphics[width=\linewidth]{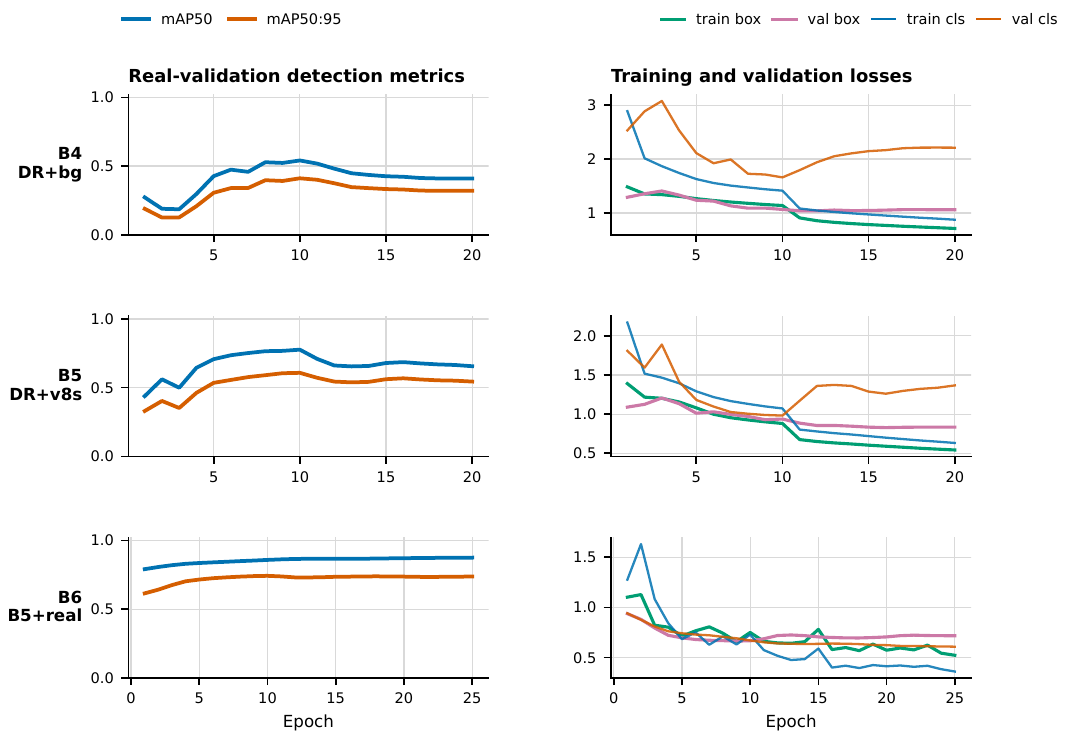}
\caption{YOLOv8~\citep{ultralytics2023yolov8} CAD-as-renderer training diagnostics for B4--B6 on T-LESS/BOP~\citep{hodan2017tless,hodan2018bop}. The same
real-validation target is used across runs. B4 adds real background negatives
to the randomized synthetic source, B5 increases detector capacity, and B6
combines the larger detector with small real calibration. The curves support
the main-text interpretation that source design, model capacity, and limited
real calibration address different parts of the CAD-rendered to real-image
domain gap.}
\label{fig:app-yolo-training-b4-b6}
\end{figure}

\begin{figure}[H]
\centering
\includegraphics[width=\linewidth,height=0.86\textheight,keepaspectratio]{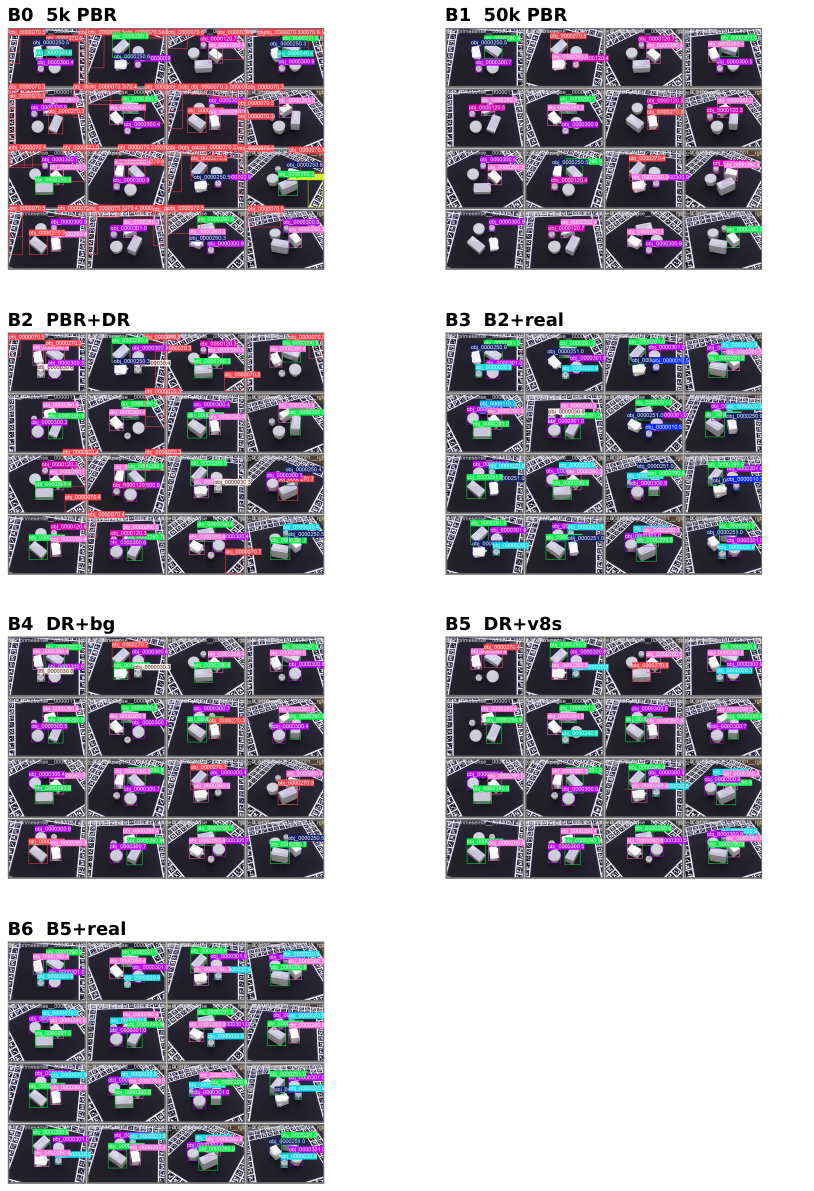}
\caption{Representative real-validation prediction grids for the seven
CAD-as-renderer detector runs. Each panel is the saved validation prediction
grid from the corresponding YOLO run on held-out real T-LESS Primesense
images~\citep{hodan2017tless}. The grids provide qualitative context for the aggregate metrics in
Figure~\ref{fig:cad-renderer-transfer}: the PBR-only models produce sparse and
unstable detections, while randomized sources, larger capacity, and small real
calibration progressively improve real-image behavior.}
\label{fig:app-yolo-validation-predictions}
\end{figure}

\begin{figure}[H]
\centering
\includegraphics[width=\linewidth,height=0.86\textheight,keepaspectratio]{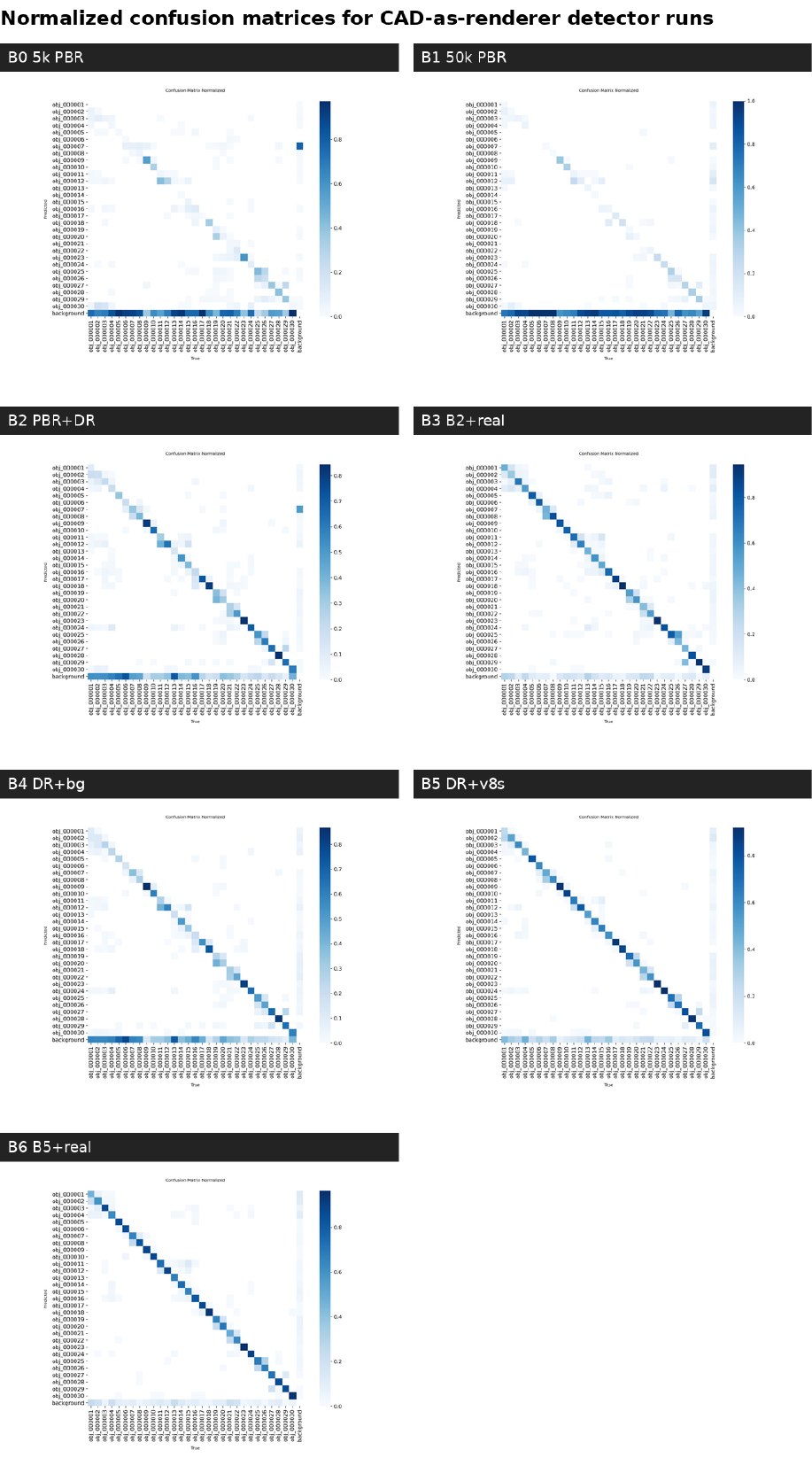}
\caption{Normalized confusion matrices for the YOLOv8~\citep{ultralytics2023yolov8} CAD-as-renderer detector runs on T-LESS~\citep{hodan2017tless}.
The matrices provide a class-level diagnostic behind the aggregate mAP scores.
They show the same progression as the main-text transfer curves: PBR-only
models have sparse and unstable real-domain class behavior, while domain
randomization, larger capacity, and small real calibration strengthen the
diagonal structure.}
\label{fig:app-yolo-confusion-matrices}
\end{figure}

\begin{figure}[H]
\centering
\includegraphics[width=\linewidth,height=0.86\textheight,keepaspectratio]{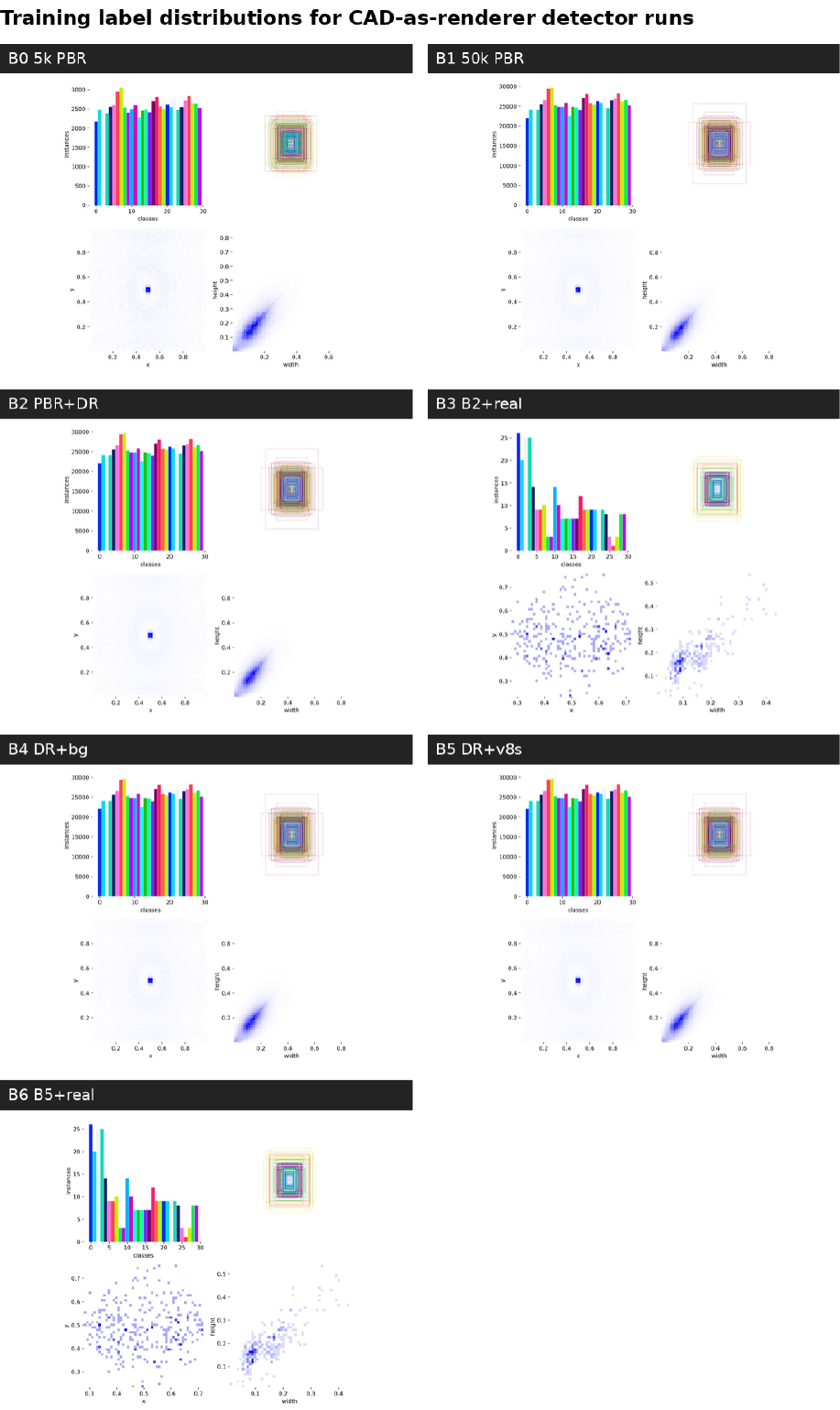}
\caption{Training label distributions for the CAD-as-renderer detector runs on T-LESS/BOP~\citep{hodan2017tless,hodan2018bop}.
The synthetic PBR and domain-randomized runs have broad synthetic class and
box distributions, while B3 and B6 show the much smaller 5\% real calibration
set. This diagnostic documents that the small-real rows in
Table~\ref{tab:cad-renderer} use real data as calibration evidence rather than
as a full replacement for synthetic rendering.}
\label{fig:app-yolo-label-distributions}
\end{figure}

\begin{figure}[H]
\centering
\includegraphics[width=\linewidth]{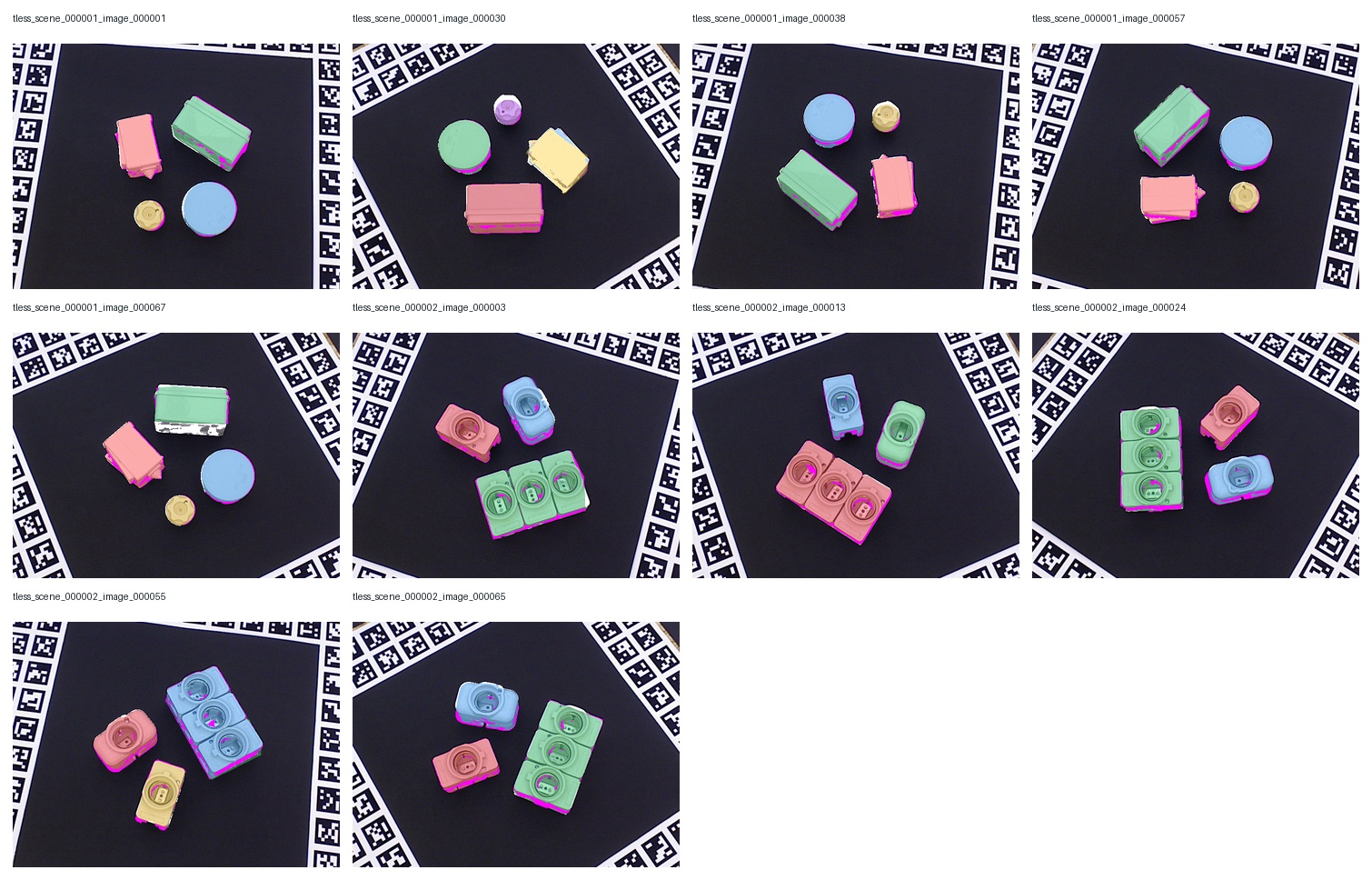}
\caption{Additional CAD-at-test-time qualitative diagnostics. The contact
sheet shows MegaPose~\citep{labbe2023megapose} render-and-compare overlays across held-out T-LESS~\citep{hodan2017tless}
Primesense scenes. It complements Figure~\ref{fig:cad-test-time-geometry} by
showing that the CAD prior is used after detection as a geometric explanation
of the observed object layout, not merely as a synthetic-image source.}
\label{fig:app-megapose-overlays}
\end{figure}

\begin{figure}[H]
\centering
\includegraphics[width=\linewidth]{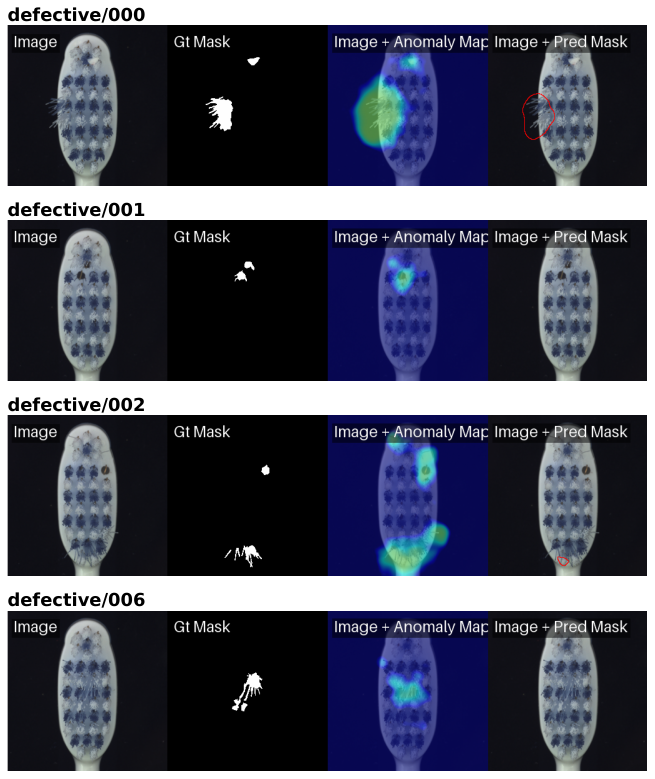}
\caption{Additional CAD-unavailable qualitative diagnostics for the synthetic-anomaly
probe. The examples show MVTec AD~\citep{bergmann2019mvtec} toothbrush test images with ground-truth
masks, anomaly heat maps, and predicted masks. This diagnostic is included as
an appearance-prior probe for the CAD-unavailable branch, rather than as a
directly comparable training curve for the detector or pose-estimation runs.}
\label{fig:app-supersimplenet-probe}
\end{figure}

\end{document}